\documentclass[11pt,letterpaper]{article}
\usepackage[T1]{fontenc}
\usepackage[utf8]{inputenc}
\usepackage{arxiv}
\usepackage{amsmath,amssymb,graphicx,booktabs,array,multirow}
\usepackage{algorithm,algpseudocode}
\usepackage{placeins}
\usepackage[numbers,sort&compress]{natbib}
\usepackage{xcolor,colortbl}
\usepackage{url}
\usepackage[hidelinks]{hyperref}
\hypersetup{pdftitle={EvoSignal: LLM-Guided Evolutionary Design of Modular Traffic Signal Control Programs},pdfauthor={Leizhen Wang and coauthors}}
\title{EvoSignal: LLM-Guided Evolutionary Design\\of Modular Traffic Signal Control Programs}
\author{Leizhen Wang$^{1,2}$\quad Peibo Duan$^{1,*}$\quad Zhenlin Qin$^{2}$\quad Yancheng Ling$^{2}$\\[3pt]
Jian Xu$^{3}$\quad Yue Wang$^{4}$\quad Hao Wang$^{5}$\quad Zhenliang Ma$^{2,*}$}
\affiliations{$^{1}$Monash University\quad $^{2}$KTH Royal Institute of Technology\\
$^{3}$Chinese Academy of Sciences\quad $^{4}$Zhejiang Dahua Technology\\
$^{5}$Southeast University\quad $^{*}$Corresponding authors}
\date{}
\begin{document}
\maketitle
\begin{abstract}
Effective traffic signal control (TSC) requires policies that respond to changing traffic demand and network conditions while meeting different control objectives. However, adapting existing strategies often involves repeated manual design and adjustment, making it difficult to systematically explore better control rules for a target network. Large language models (LLMs) can automate this process, but directly using them to select signal phases leaves decision rules embedded in black-box models and incurs recurring inference costs and latency. This paper formulates TSC as a modular program design problem and proposes EvoSignal, an LLM-guided evolutionary framework using traffic knowledge and performance feedback. The modular representation separates traffic feature extraction, local phase prioritization, and optional network-based priority adjustment. Starting from several established strategies, EvoSignal improves programs through feedback on congestion and signal operation, retaining strategies with different performance trade-offs. The resulting programs operate without online LLM inference. Simulation experiments across five scenarios on two real-world road networks show that the selected default EvoSignal program reduces waiting time by 16.8--49.2\% relative to the lowest waiting time achieved by the 12 conventional, reinforcement learning-based, and LLM-based baselines in each scenario. A program prioritizing travel time and queue length outperforms all 12 baselines on all three metrics in the search scenario and remains among the top two on each metric when transferred unchanged to the other four scenarios. These findings support automated design of inspectable control programs that transfer across the evaluated road networks and traffic demands.
Code is available at \url{https://github.com/georgewanglz2019/EvoSignal}.
\end{abstract}

\section{Introduction}
\label{sec:introduction}
Traffic signal control (TSC) is an important component of urban traffic management. By selecting signal phases that serve competing traffic movements, a controller can reduce vehicle queues and waiting times at intersections. Effective control must account for changing traffic demand and interactions between intersections: releasing vehicles from one intersection may increase congestion on downstream roads and affect traffic conditions elsewhere in the network. These characteristics make the design of effective control policies challenging, particularly when road networks, traffic demand, and control objectives differ across scenarios~\citep{papageorgiou2003review,chen2022real}.

Existing TSC methods include conventional traffic engineering approaches and reinforcement learning (RL)-based approaches. Conventional methods, such as fixed-time control, adaptive signal control, and max-pressure control, provide well-established principles for determining signal settings~\citep{webster1958traffic,hunt1982scoot,varaiya2013max}. Their control rules can respond to traffic conditions, but selecting and extending these rules for a target network still requires engineering effort. RL-based methods learn policies through interactions with a traffic environment and have achieved strong performance in signal coordination and network-wide control~\citep{wei2019presslight,wei2019colight,chen2020toward}. However, training requires interaction data and computation, and neural policies can be difficult to inspect and modify when control requirements change~\citep{chen2022real}. Programmatic policies offer explicit decision rules that can be examined and adjusted~\citep{gu2024pi}, motivating automated design methods that retain this form of control.

Recent studies use large language models (LLMs) to bring traffic knowledge and reasoning into signal control. LLMLight and Traffic-R1 analyze traffic conditions and select signal phases through LLM inference~\citep{lai2025llmlight,zou2025traffic}. This places language generation within the repeated control loop, adding inference latency and computational demand across intersections. Traffic-specialized models, including LightGPT and Traffic-R1, also require data collection and model training, although smaller models and efficient inference can mitigate deployment costs~\citep{lai2025llmlight,zou2025traffic}. Their language-based rationales help explain individual actions, but the complete decision rule remains encoded in model parameters and cannot be directly inspected and edited as a control program. These considerations motivate using LLM capabilities to improve controller design while keeping online phase selection computationally simple and explicit.

An alternative is to use an LLM to generate and revise a control program offline. The resulting program selects phases directly from traffic observations, removing repeated LLM inference from operation and exposing the implemented decision rules for inspection. Program revision can also adapt the controller without updating the LLM's parameters. LLM-guided program evolution provides a mechanism for this process: candidate code is generated, evaluated, and improved using feedback. FunSearch, AlphaEvolve, OpenEvolve, and ShinkaEvolve establish this approach for program optimization~\citep{romera2024mathematical,novikov2025alphaevolve,openevolve,lange2025shinka}. Their general search mechanisms leave the traffic control representation and domain-specific evaluation guidance to the application designer. In TSC, the earlier EvolveSignal study explores fixed-time strategy design at a single intersection~\citep{wang2025evolvesignal}, while SignalClaw evolves interpretable skills using structured traffic feedback and event-based composition~\citep{lei2026signalclaw}. Building on these advances, an important question is how to organize the evolution of state-responsive control programs across a road network, using existing strategies and observed traffic behavior to meet different control objectives.

This design problem involves four connected challenges. First, control decisions depend on local queues and approaching vehicles as well as downstream and network conditions. Evolving an undifferentiated program makes it harder to revise these functions separately or match them to available observations. Second, a single initial program represents only one control strategy, leaving other established approaches to pressure, demand, and service fairness unused at initialization. Third, aggregate performance scores rank programs but do not identify where queues accumulate or which phase-selection patterns contribute to long waiting times. Effective revision requires diagnostic feedback that connects observed traffic behavior to control logic. Finally, travel time, queue length, and waiting time can favor different strategies. Selecting only by a combined score can overlook useful alternatives and their reusable control ideas. These challenges call for a traffic-specific organization of program structure, initialization, diagnostic feedback, and strategy reuse within the evolutionary process.

We formulate TSC as a modular program design problem: an executable policy maps traffic observations to signal phase decisions, and its control logic is optimized for specified performance objectives. To address this problem, we propose EvoSignal, an LLM-guided evolutionary framework. Its three-module representation (3M) separates traffic feature extraction, local phase prioritization, and optional network-based priority adjustment, allowing control logic to reflect the available traffic information. Multi-strategy initialization (Multi-init) supplies programs based on established control strategies and complementary design principles. The simulation performance analyzer (SPA) diagnoses congestion and phase-selection behavior and suggests program revisions, while a Pareto strategy archive (PSA) supplies experience from recently evaluated strategies with different performance trade-offs. The LLM uses these inputs to revise candidate programs, whose measured control performance guides further evolution. Feedback consists of traffic observations and performance measurements; in this study, these are obtained through simulation. The selected program subsequently determines signal phases directly from traffic observations.

\begin{samepage}
The main contributions are summarized as follows:
\begin{enumerate}
\item We formulate TSC as a modular program design problem, where an executable policy maps traffic observations to signal phase decisions under specified control objectives. The representation connects traffic feature extraction, local phase prioritization, and optional network-based priority adjustment through explicit interfaces, enabling targeted revision and inspection of the control logic.
\item We develop EvoSignal, an evolutionary framework that combines multiple initial control strategies, diagnostic feedback on traffic control performance, and reuse of strategies with different objective trade-offs to guide LLM-based program revision.
\item We evaluate EvoSignal across five scenarios on two real-world road networks against conventional, RL-based, and LLM-based controllers, and LLM-guided program evolution frameworks. The results demonstrate competitive control performance and more consistent search outcomes. A program evolved in one scenario transfers unchanged to the other four scenarios, ranking among the top two on travel time, queue length, and waiting time against 12 reference controllers.
\end{enumerate}
\end{samepage}

The remainder of this paper is organized as follows. Section~\ref{sec:review} reviews related work. Section~\ref{sec:method} formulates the modular program design problem and presents the EvoSignal search framework. Section~\ref{sec:experiment} describes the experimental setup and evaluates control performance, search behavior, and component contributions. Section~\ref{sec:discussion} analyzes the evolved programs and discusses their implications and limitations. Section~\ref{sec:conclusion} concludes the paper.

\section{Related Work}
\label{sec:review}
We review four groups of methods: conventional TSC, RL-based TSC, LLM-based TSC, and LLM-guided program design and evolution. The discussion follows how these methods construct and execute control policies, emphasizing their ability to respond to traffic conditions, the transparency of their decision rules, and their computational requirements. Table~\ref{tab:tsc-approach-comparison} summarizes these representative formulations in terms of policy learning, explicit program revision, and online decision making.

\begin{table}[!htbp]
\centering
\footnotesize
\caption{Comparison of representative TSC formulations, with EvoSignal illustrating LLM-guided program design and evolution.}
\label{tab:tsc-approach-comparison}
\setlength{\tabcolsep}{3pt}
\renewcommand{\arraystretch}{1.25}
\begin{tabular}{@{}>{\raggedright\arraybackslash}p{0.39\linewidth}*{3}{>{\centering\arraybackslash}p{0.13\linewidth}}>{\centering\arraybackslash}p{0.16\linewidth}@{}}
\toprule
\textbf{Property} & \shortstack{Conventional\\TSC\\\citep{varaiya2013max}} & \shortstack{RL-based TSC\\(neural policies)\\\citep{wei2019presslight,wei2019colight}} & \shortstack{LLM-based\\TSC\\\citep{lai2025llmlight,zou2025traffic}} & \shortstack{LLM-guided\\program design\\and evolution\\(EvoSignal)} \\
\midrule
Learns adaptive control policies from experience or feedback & $\times$ & $\checkmark$ & $\checkmark$ & $\checkmark$ \\
Uses pretrained LLM knowledge in design or decision making & $\times$ & $\times$ & $\checkmark$ & $\checkmark$ \\
Supports online reasoning with natural-language rationales & $\times$ & $\times$ & $\checkmark$ & $\times$ \\
Automatically revises explicit traffic-feature computations & $\times$ & $\times$ & $\times$ & $\checkmark$ \\
Automatically revises explicit control rules & $\times$ & $\times$ & $\times$ & $\checkmark$ \\
Exposes executable decision logic for human inspection and editing & $\checkmark$ & $\times$ & $\times$ & $\checkmark$ \\
No online neural-model inference (including LLMs) & $\checkmark$ & $\times$ & $\times$ & $\checkmark$ \\
\bottomrule
\end{tabular}
\par\smallskip
\begin{minipage}{\linewidth}
\scriptsize
$\checkmark$: Supported; $\times$: Not part of the representative formulation.
\par\smallskip
Conventional controllers can respond to traffic changes without learning new policies. The RL column covers neural policies; the LLM column includes traffic-trained models such as LightGPT and Traffic-R1. Feature revision concerns explicit code rather than latent representations. Learning occurs during training or program design; online denotes phase selection during deployment.
\end{minipage}
\end{table}

\subsection{Conventional Traffic Signal Control}
Conventional TSC methods determine signal settings using traffic engineering models and predefined control rules~\citep{papageorgiou2003review}. Fixed-time control allocates green time according to estimated demand; Webster's formulation provides a classical basis for cycle length and green-split design~\citep{webster1958traffic}. Adaptive systems such as SCOOT adjust signal settings using measured traffic conditions~\citep{hunt1982scoot}. Their implementation must account for phase structure, detector availability, and timing constraints~\citep{koonce2008manual}.

Pressure-based control uses queue information to determine which movements to serve. Max-pressure control selects phases according to upstream--downstream queue differentials and provides network stability guarantees under its modeling assumptions~\citep{varaiya2013max}. Capacity-aware backpressure accounts for limited downstream storage~\citep{gregoire2015capacity}, while cyclical max-pressure control incorporates an operational phase structure~\citep{levin2020cyclical}. Distributed adaptive control also uses shockwave-based delay estimates and downstream queues to guide phase activation~\citep{xie2025shockwave}. Conventional methods offer explicit rules, modest computational requirements, and, for certain pressure-based policies, theoretical stability guarantees. Adaptive and pressure-based controllers already respond to changing traffic conditions, but their responses remain constrained by predefined control rules. Extending these rules to capture more complex traffic interactions or changing objectives requires further engineering effort. This motivates learning control policies from traffic experience.

\subsection{Reinforcement Learning-Based Traffic Signal Control}
RL learns decision policies through interaction with an environment, with transportation applications including traffic assignment~\citep{wang2026sequential,wang2025assignment}, and adaptive signal control~\citep{wei2021recent,haydari2022deep}. For TSC, this addresses the rule-design limitation by learning how signal actions should depend on traffic conditions. Early work investigates adaptive intersection control~\citep{abdulhai2003reinforcement}, multi-agent coordination~\citep{arel2010reinforcement}, and integrated control over large road networks through MARLIN-ATSC~\citep{eltantawy2013multiagent}. Deep RL extends these approaches by learning policies from richer traffic observations~\citep{li2016traffic,wei2018intellilight}.

A central design choice is how to represent traffic demand and compare competing phases. FRAP models phase competition~\citep{zheng2019frap}, PressLight incorporates pressure into policy learning~\citep{wei2019presslight}, and MPLight supports shared decentralized control across intersections~\citep{chen2020toward}. AttendLight uses lane and phase attention to accommodate different intersection configurations~\citep{oroojlooy2020attendlight}. Efficient-CoLight uses efficient pressure to represent traffic movements~\citep{wu2021efficient}, while Advanced-CoLight also considers approaching vehicles~\citep{zhang2022expression}. TransformerLight captures temporal traffic patterns through sequence modeling~\citep{wu2023transformerlight}. Control objectives extend beyond vehicle efficiency to person-based waiting costs across transport modes~\citep{wang2023hmd}, while action designs include joint phase selection and duration optimization~\citep{jin2026hybrid} and cycle-level optimization of cycle length and green splits~\citep{han2024cyclight}.

Network coordination and adaptation introduce further challenges. Attention mechanisms support network coordination through graph-based modeling of neighboring intersections~\citep{wei2019colight} and the selection of relevant intersection information for centralized critics~\citep{han2024attention}. Multi-agent actor--critic control~\citep{chu2020multiagent} and knowledge sharing in KS-DDPG~\citep{li2021network} support cooperation, while MetaLight reuses experience to adapt to new scenarios~\citep{zang2020metalight}, and adversarial training improves robustness to environmental mismatch in integrated expressway and intersection control~\citep{han2024adversarial}. Reference-guided action selection also incorporates established timing policies to improve early training performance~\citep{lu2025reference}. Programmatically interpretable RL~\citep{verma2018programmatically} and $\pi$-Light~\citep{gu2024pi} further address policy transparency using explicit programs. Overall, RL expands the ability to learn traffic-responsive behavior and coordination, but requires interaction data and optimization. For neural controllers, inspecting or modifying a particular decision rule remains difficult. Generalization beyond the training conditions is also a continuing challenge, and unfamiliar networks or traffic patterns may require additional adaptation~\citep{wei2021recent,chen2022real}. These limitations motivate incorporating broader traffic knowledge and more accessible explanations into control decisions.

\subsection{LLM-Based Traffic Signal Control}
LLMs are increasingly applied to transportation prediction, information analysis, and decision support~\citep{ma2026urban}. These applications use traffic context and travel histories for mobility prediction~\citep{wang2023llmmob,qin2025lingotrip} and day-to-day route choice~\citep{wang2025agentic}, or connect language interfaces to traffic analysis tools~\citep{zhang2024trafficgpt}. In TSC, pretrained knowledge and language reasoning support the interpretation of traffic conditions and signal decisions. Here, the LLM acts as the decision maker: traffic observations are expressed in a prompt, and model inference determines the signal action. LLMLight applies this approach to phase selection and develops LightGPT through traffic-specific training~\citep{lai2025llmlight}. Our experiments include LLMLight with GPT-4 and LightGPT with Llama2-13B. The GPT, Llama, and Qwen families provide representative backbones~\citep{openai2023gpt4,touvron2023llama2,grattafiori2024llama3,yang2024qwen2}. More generally, chain-of-thought prompting~\citep{wei2022chain} and self-consistency~\citep{wang2023selfconsistency} study how language models can reason through decisions and combine multiple reasoning paths.

Domain adaptation has been investigated for transferable mobility prediction~\citep{qin2026mobility}, information extraction from crash narratives~\citep{wang2026crash}, and traffic-specific phase decisions~\citep{lai2025llmlight,zou2025traffic}. Within TSC, methods also extend the information and learning mechanisms available to the LLM decision maker. LLM-Assisted Light places the LLM at the center of control and supplies perception and decision tools~\citep{wang2024assisted}. Traffic-R1 trains traffic reasoning through RL and supports communication between intersections~\citep{zou2025traffic}. These developments improve the use of traffic knowledge, support transfer to unfamiliar conditions, and make individual decisions easier to discuss in language. Nevertheless, repeated model inference introduces computational cost and latency into online control. Compact models reduce this burden~\citep{zou2025traffic}, but resource requirements remain a deployment consideration. A verbal rationale alone does not expose the complete neural decision rule or establish that it faithfully accounts for the selected action, limiting direct inspection and modification. These considerations motivate using LLMs to design executable control programs offline, so that online phase selection can use explicit code without repeated language-model inference.

\subsection{LLM-Guided Program Design and Evolution}
This approach separates the use of LLM reasoning from online control execution. Code language models can generate executable programs from task descriptions~\citep{chen2021code}, and AlphaCode demonstrates large-scale generation and filtering of candidate solutions~\citep{li2022competition}. LLM-guided program evolution further combines code generation with repeated performance evaluation. FunSearch retains and improves promising programs using computational feedback~\citep{romera2024mathematical}, while AlphaEvolve develops an evolutionary coding workflow around automated evaluators~\citep{novikov2025alphaevolve}. OpenEvolve provides the open-source infrastructure used in this study~\citep{openevolve}. ShinkaEvolve improves search through parent sampling, novelty-based filtering, and adaptive model selection, while retaining metrics and textual feedback for subsequent revisions~\citep{lange2025shinka}.

Related approaches differ in the programs they produce and the feedback they use. Evolution of Heuristics jointly improves descriptions and code for optimization rules~\citep{liu2024eoh}, while Eureka generates reward functions for training RL policies~\citep{ma2024eureka}. Voyager stores executable skills and refines them using environmental feedback~\citep{wang2023voyager}. Self-Refine iterates between feedback and revision~\citep{madaan2023selfrefine}, Reflexion retains verbal reflections from task outcomes~\citep{shinn2023reflexion}, and ReEvo incorporates reflection into LLM-based evolutionary search~\citep{ye2024reevo}. These studies motivate using both measured performance and textual feedback to guide code improvement without updating the LLM's parameters.

For traffic signals, the earlier EvolveSignal study applies a coding agent to fixed-time strategy design at a single intersection~\citep{wang2025evolvesignal}. SignalClaw evolves interpretable traffic skills using structured feedback and event-based composition~\citep{lei2026signalclaw}. These approaches demonstrate the value of evaluating and revising executable policies. However, general program-evolution frameworks leave the organization of traffic control logic and domain-specific guidance to the application designer. Traffic-specific studies address particular control settings and skill structures, leaving room to investigate how local decisions and network conditions can be combined with multiple existing strategies and different control objectives. EvoSignal addresses this design question through modular control programs, multiple initial strategies, traffic diagnostics, and reuse of strategies with complementary performance trade-offs. The resulting rules can be inspected and executed without online LLM inference, while their effectiveness and transfer still require evaluation across the intended traffic conditions.

\section{Methodology}
\label{sec:method}

\subsection{TSC as a Modular Program Design Problem}
\label{sec:problem}
We formulate TSC policy design as the search for an executable program that maps traffic observations to signal phase decisions. We first define the control task and phase-priority-based decision rule, then specify the modular program representation and its performance objective.

\subsubsection{Traffic observations and phase decisions}
Consider a road network with signalized intersections. At decision step $t$, intersection $i$ selects a phase from its admissible phase set $P_i$. A phase serves a group of non-conflicting traffic movements, following the phase-selection setting used in LLMLight~\citep{lai2025llmlight}. The input $x_{i,t}$ includes the current phase, elapsed phase time, vehicle counts and queues by traffic movement, and available downstream traffic information. It can also include decision history and information shared by other intersections.

An executable control program $p$ maps this input to the selected phase $a_{i,t}$:
\begin{equation}
a_{i,t}=p(x_{i,t})\in P_i.
\label{eq:policy}
\end{equation}
The same program runs at different intersections using their respective observations and state. Its code remains fixed during control evaluation, while its decisions respond to changing traffic conditions.

We adopt \emph{phase-priority-based control}: at each decision step, the program computes a priority $s_k(x_{i,t})$ for each admissible phase $k$ and selects
\begin{equation}
a_{i,t}\in\arg\max_{k\in P_i}s_k(x_{i,t}).
\label{eq:phase-selection}
\end{equation}
Ties are resolved by the first phase in the configured ordering. Selecting the current phase again extends service to the same movements; selecting another phase invokes the environment's signal-transition procedure. During a transition, the controller preserves its previous action. Section~\ref{sec:setup} specifies the phase configuration and timing settings used in the experiments.

This decision structure connects traffic observations to explicit phase preferences. Recomputing priorities allows the service order to respond to changing demand. For automated program design, the structure makes competing phase preferences available for inspection and permits the scoring logic to be revised while retaining a simple selection operation. It also accommodates pressure, approaching demand, and service history within a common control interface. EvoSignal organizes these computations through a three-module representation (3M).

\subsubsection{3M: Modular control program representation}
\label{sec:3m}
Traffic signal control requires a controller to interpret observations, prioritize competing movements, and consider surrounding traffic conditions. The 3M representation separates these tasks into traffic feature extraction (M1), local phase prioritization (M2), and network-based priority adjustment (M3). A candidate program contains the three functions $F_1,F_2,F_3$; their input--output interfaces are specified, while their internal computations are revised during evolution.

\begin{figure}[!htbp]
\centering
\includegraphics[width=\linewidth]{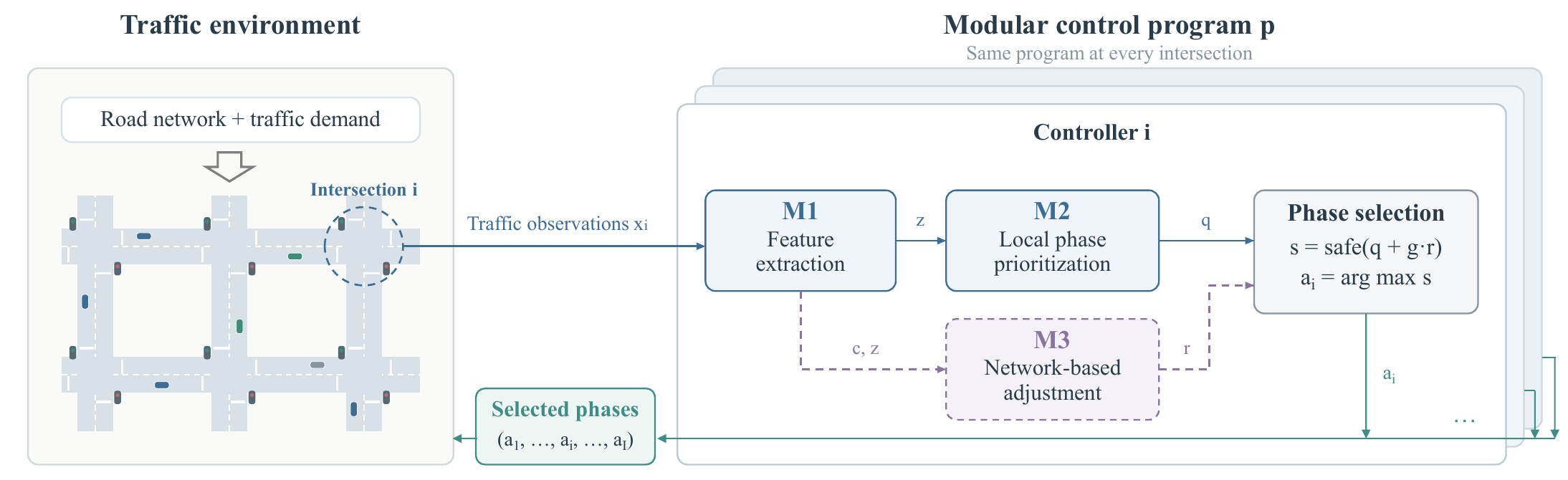}
\caption{Modular control program at decision step $t$. One controller is instantiated per intersection, with controller $i$ expanded; the selected phases jointly control the traffic environment.}
\label{fig:controller-execution}
\end{figure}

Figure~\ref{fig:controller-execution} shows execution at decision step $t$. Each of the $I$ intersections runs the same program using its own observations and state; the selected phases $(a_{1,t},\ldots,a_{I,t})$ jointly control the traffic environment, whose next observations continue the loop. Program code stays fixed during execution; Figure~\ref{fig:evosignal-framework} shows how it is revised during design.

For one intersection at one decision step, omit the intersection and time subscripts and let $x$ denote the available observations, history, and shared state. The main computations are
\begin{align}
(z,c)&=F_1(x),\label{eq:m1}\\
q&=F_2(z),\label{eq:m2}\\
r&=F_3(c,z).\label{eq:m3}
\end{align}
Here $z$ contains features for phase prioritization, $c$ contains additional traffic context, and $q,r\in\mathbb{R}^{|P|}$ give the base scores and adjustments for the admissible phases in $P$. Figure~\ref{fig:controller-execution} omits $t$ for readability: $x_i,a_i$ denote $x_{i,t},a_{i,t}$, while its remaining variables follow these interfaces and Eq.~\eqref{eq:action}. These equations specify the module interfaces. The functions may also use controller history and shared state; the features and scoring expressions within each function can evolve.

\paragraph{M1: Traffic feature extraction}
M1 transforms movement-level observations into features for phase selection, such as pressure, waiting and approaching demand, downstream congestion, and service history. These form $z$, while $c$ can include adjacency information, shared features, and network summaries. Evolution can revise the feature definitions and combinations. Appendix~\ref{sec:initial-program-details} gives a concrete pressure--demand example.

\paragraph{M2: Local phase prioritization}
M2 assigns a base score $q_k$ to each phase using the extracted features. Evolution can revise coefficients and introduce nonlinear, conditional, or history-dependent rules; the interface does not restrict candidates to a fixed scoring expression.

\paragraph{M3: Network-based priority adjustment}
M3 computes $r_k$ from the traffic context and extracted features. Its role is to adjust a local phase preference in response to conditions beyond the current intersection. A gate $g$ controls the contribution of M3. The final phase priorities are
\begin{equation}
s_k=\operatorname{safe}(q_k+g r_k),\qquad k\in P,
\label{eq:action}
\end{equation}
where $\operatorname{safe}$ clips scores to the configured range and replaces nonfinite values before selection. The evolved program may adjust $g$. At non-transition decision steps, these priorities determine the phase through Eq.~\eqref{eq:phase-selection}. Together, the three functions and the phase-selection operation implement the executable policy in Eq.~\eqref{eq:policy}.

The three stages make the information requirements and decision components explicit. A controller can use features supported by its available observations and include an additional network-based adjustment when appropriate. The generated code determines its actual dependencies. Section~\ref{sec:selected-rules} compares two selected programs, with detailed feature extraction and priority rules in Appendix~\ref{sec:selected-programs}.

\subsubsection{Program design objective}
Let $\Pi$ denote the class of executable control programs that follow the 3M interfaces and return admissible phase decisions. We evaluate control performance using travel time (TT), queue length (QL), and waiting time (WT), all of which are minimized. For a given road network, traffic demand, and evaluation horizon, the performance of a program $p\in\Pi$ is represented by
\begin{equation}
\mathbf{f}(p)=\bigl(f_T(p),f_Q(p),f_W(p)\bigr)
=(\mathrm{TT},\mathrm{QL},\mathrm{WT}),
\label{eq:objectives}
\end{equation}
TT averages the accumulated entry-to-exit durations in recorded vehicle trajectories. QL is the time average of the total number of queued vehicles on network approaches. WT is the time average of the mean elapsed duration of active stopping episodes at each sample, with zero assigned to an empty sample.

To express the control objectives during program design, we use the bounded scalar score
\begin{equation}
S(p)=\sum_{j\in\{T,Q,W\}}w_j
\frac{1}{1+f_j(p)/C_j},
\label{eq:fitness}
\end{equation}
where $C_j>0$ is a metric-specific scale. The weights satisfy $w_j\geq0$ and $\sum_j w_j=1$ and remain fixed within a search run. A larger score indicates better performance under the chosen weights. Section~\ref{sec:setup} specifies the experimental weights, and Appendix~\ref{sec:implementation-settings} lists the metric scales.

The modular program design problem is
\begin{equation}
\max_{p\in\Pi} S(p).
\label{eq:design-objective}
\end{equation}
The design variables are the executable implementations of $F_1,F_2,F_3$, including their feature computations, scoring expressions, conditional logic, and numerical coefficients. The module interfaces and the phase-selection operation remain specified. The priority $s_k$ determines which phase to serve at one decision step, whereas $S(p)$ evaluates the traffic consequences of running the program over the evaluation horizon. Improving a program therefore requires evaluating its successive decisions across the network. Each search uses a specified scenario and fixed objective weights; reuse under other networks and demands is evaluated separately.

\subsection{EvoSignal: LLM-Guided Evolutionary Search}
\label{sec:overview}
\begin{figure}[!htbp]
\centering
\includegraphics[width=\linewidth]{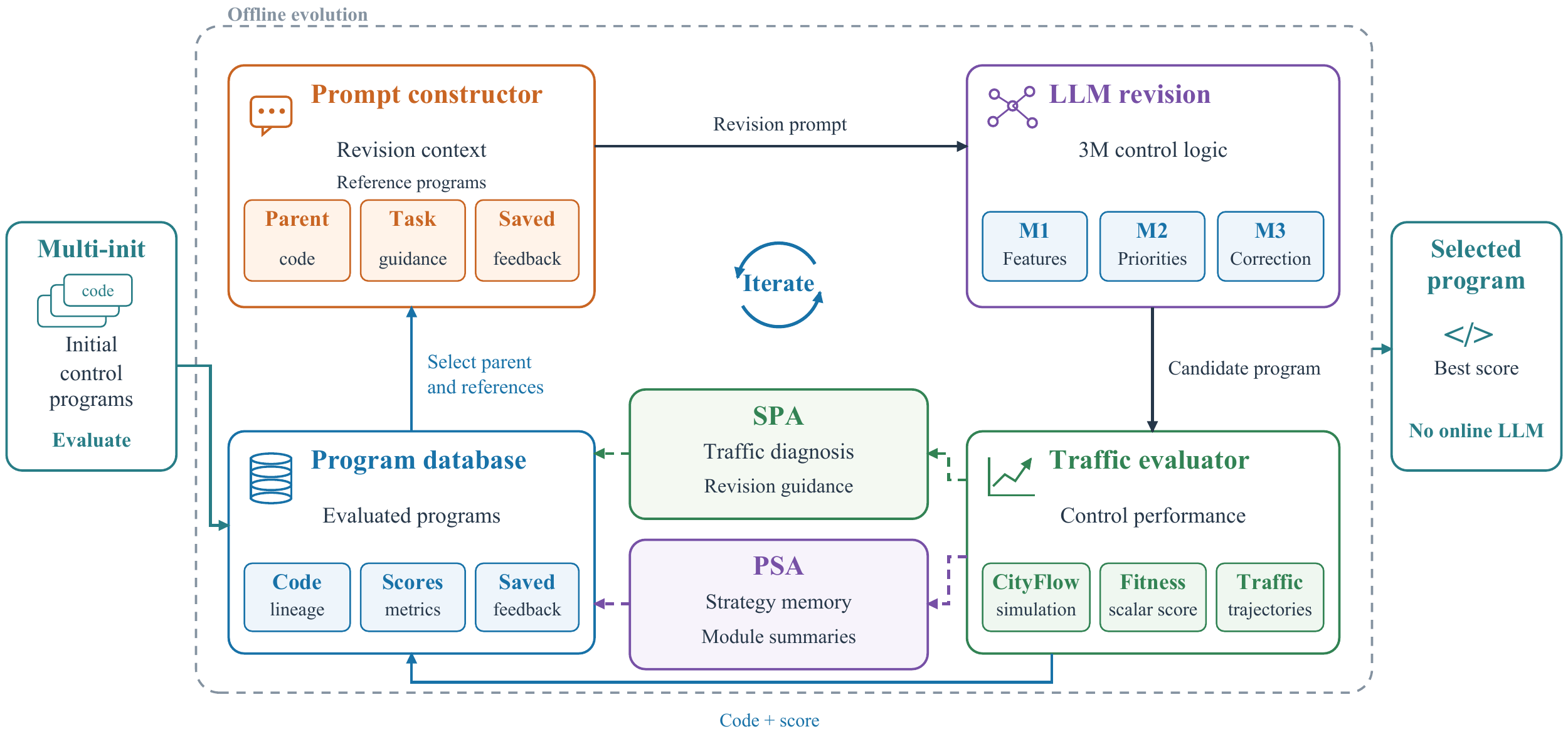}
\caption{EvoSignal framework for evolving modular control programs from multiple initial strategies, guided by traffic diagnostics and reusable strategy summaries. The selected program controls signals without online LLM inference.}
\label{fig:evosignal-framework}
\end{figure}

EvoSignal addresses Eq.~\eqref{eq:design-objective} through the revision--evaluation loop in Figure~\ref{fig:evosignal-framework}. Multi-init supplies initial programs; the program database selects parents and references; and the LLM revises control logic within the 3M interfaces. The traffic evaluator measures each candidate's performance, SPA diagnoses its traffic behavior, and PSA supplies summaries of strategies with different performance trade-offs. These records guide subsequent revisions. EvoSignal builds on the program database and code-editing mechanisms of OpenEvolve v0.2.26~\citep{openevolve}, a community-developed open-source implementation of AlphaEvolve~\citep{novikov2025alphaevolve}. In this study, traffic feedback is obtained through simulation.

\subsubsection{Evolution procedure}
Algorithm~\ref{alg:evosignal} summarizes the procedure. The database $D$ stores code, parent identities, execution status, and evaluation records. A successful program has record $m=(\mathbf{f},S,d,b)$: traffic metrics, combined score, SPA feedback, and a saved PSA summary. The inherited database uses islands, elite retention, and exploratory sampling to select a parent $p_a$ and reference programs $R_n$. The guidance $G_n$ specifies the task, interfaces, and current revision emphasis. The prompt constructor $H$ combines these inputs, and the fixed-parameter LLM distribution $P_\theta$ generates edits. PSA memory $M$ retains the latest $L$ strategy records. Appendix~\ref{sec:evolution-details} gives storage and update details.

Let $V_N$ contain all valid evaluated programs from initialization and $N$ evolution attempts. EvoSignal returns the best observed program:
\begin{equation}
p^\star\in\arg\max_{p\in V_N}S(p).
\label{eq:program-selection}
\end{equation}
Selection uses the source scenario and does not assume a global optimum over $\Pi$; the selected code remains fixed during control execution.

\begin{algorithm}[H]
\caption{EvoSignal evolutionary design procedure}
\label{alg:evosignal}
\footnotesize
\begin{algorithmic}[1]
\Require Initial programs $p^{(1)},\ldots,p^{(K)}$; traffic scenario; weights $w_j$; guidance $G$; budget $N$
\Ensure Selected executable program $p^\star$
\State Initialize database $D$ and PSA memory $M$
\State Evaluate initial programs; store their code, metrics, scores, and status in $D$
\State Summarize successful initial programs and update $M$
\For{$n=1,\ldots,N$}
    \State Select parent $p_a$ with record $m_a$ and references $R_n$ from $D$
    \State Construct prompt $u_n=H(p_a,m_a,R_n,G_n)$
    \State Generate $y_n\sim P_\theta(\cdot\mid u_n)$; apply edits $p_n=\operatorname{Edit}(p_a,y_n)$
    \State Check and evaluate $p_n$; record failures and continue if unsuccessful
    \State Obtain metrics $\mathbf{f}(p_n)$ and trajectory $\tau_n$; compute $S(p_n)$
    \State Construct SPA feedback $d_n$ from the trajectory and metrics
    \State Summarize the modules, update $M$, and construct PSA summary $b_n$
    \State Store $p_n$, its parent identity, and record $m_n=(\mathbf{f}(p_n),S(p_n),d_n,b_n)$ in $D$
\EndFor
\State \Return $p^\star$ selected by Eq.~\eqref{eq:program-selection}
\end{algorithmic}
\end{algorithm}

For each revision, the prompt combines control code with measured outcomes and saved feedback:
\begin{equation}
u_n=H(p_a,m_a,R_n,G_n),
\label{eq:revision-prompt}
\end{equation}
It includes the parent's saved PSA summary and can emphasize different modules at different search stages without restricting edits to those modules. The LLM then proposes changes:
\begin{equation}
y_n\sim P_{\theta}(\cdot\mid u_n),
\qquad p_n=\operatorname{Edit}(p_a,y_n).
\label{eq:revision-proposal}
\end{equation}
Here $\operatorname{Edit}$ applies SEARCH/REPLACE instructions (Appendix~\ref{sec:case-output}), while the LLM parameters $\theta$ remain fixed. Successful evaluation yields traffic metrics and a trajectory $\tau_n$ containing observations, phase decisions, and module outputs. SPA and PSA convert these outcomes into feedback stored with the candidate for later revision. Appendix~\ref{sec:revision-example} shows the actual prompt, response, code edits, and measured outcome of one iteration.

\subsection{Multi-init: Initialization with Multiple Control Strategies}
\label{sec:multi-init}
Multi-strategy initialization (Multi-init) introduces established control knowledge and complementary design principles into the initial population. Seeds can adapt traffic-control methods such as max-pressure or use LLM-generated implementations inspired by feedback control, fair service, and spatial coordination. Each expresses a control idea through the same 3M interfaces. These starting points are intended to diversify decision rules, reduce dependence on a single strategy, and supply alternative ideas for later revisions; the framework does not prescribe their number or types.

Let $p^{(1)},\ldots,p^{(K)}$ denote the seeds. Each is evaluated under the same scenario and objective weights:
\begin{equation}
\bigl(\mathbf{f}(p^{(k)}),\tau^{(k)}\bigr)=E(p^{(k)}),
\qquad k=1,\ldots,K.
\label{eq:seed-evaluation}
\end{equation}
The evaluator $E$ returns traffic metrics and a trajectory for each successful program. The evaluated seeds initialize $D_0$, from which parents and references are selected. Seed evaluations precede the $N$ evolution attempts. Appendices~\ref{sec:initial-program-details} and~\ref{sec:evolution-details} describe the experimental seeds and their initialization records.

\subsection{SPA: Simulation Performance Analyzer}
\label{sec:spa}
The scalar score ranks overall performance but does not show where queues accumulate or how phase selection relates to demand. SPA converts recorded traffic behavior into diagnostics and revision suggestions using deterministic statistics and predefined rules. The LLM uses this report together with the program code to propose changes.

For trajectory $\tau_n$, let $I$ be the number of intersections, $T$ the number of diagnostic samples, $Q_{i,t,m}$ the waiting-vehicle count for movement $m$, and $a_{i,t}$ the recorded phase action. With $B_k$ denoting movements served by phase $k$, SPA relates average queued demand to selection frequency:
\begin{equation}
\begin{aligned}
\overline{Q}^{\mathrm{phase}}_k&=\frac{1}{IT}\sum_{i=1}^{I}\sum_{t=1}^{T}\sum_{m\in B_k}Q_{i,t,m},\\
v_k&=\frac{1}{IT}\sum_{i=1}^{I}\sum_{t=1}^{T}\mathbf{1}(a_{i,t}=k).
\end{aligned}
\label{eq:spa-service}
\end{equation}
Here $\mathbf{1}(\cdot)$ is the indicator function. The pair $(\overline{Q}^{\mathrm{phase}}_k,v_k)$ helps identify mismatches between demand and phase service, although selection frequency alone does not establish whether green time is adequate. SPA also examines switching, queue changes, and congestion across locations and time windows; Appendix~\ref{sec:diagnostic-details} gives these statistics and diagnostic settings.

The feedback is
\begin{equation}
d_n=\operatorname{SPA}\bigl(\tau_n,\mathbf{f}(p_n)\bigr),
\label{eq:spa-feedback}
\end{equation}
The report suggests directions such as checking phase-continuation rules, demand--priority alignment, or service under late congestion. Subsequent evaluation determines whether an edit improves performance. Feedback $d_n$ is stored with the candidate and enters Eq.~\eqref{eq:revision-prompt} when it becomes a parent. Appendix~\ref{sec:case-spa} shows a report and its corresponding revision.

\subsection{PSA: Pareto Strategy Archive}
\label{sec:psa}
A program with a lower aggregate score may contain useful rules for reducing queues or waiting. PSA retains module summaries and measured outcomes from recent successful evaluations so that such control ideas can inform later revisions. An auxiliary LLM call summarizes each program's three modules as $s_n=(s_n^1,s_n^2,s_n^3)$. The updated memory $M^+$ retains the latest $L$ records; it includes successful seed evaluations.

All three traffic metrics are minimized. Program $p_a$ dominates $p_b$ when
\begin{equation}
\begin{aligned}
f_j(p_a)&\leq f_j(p_b) &&\text{for all }j\in\{T,Q,W\},\\
f_j(p_a)&<f_j(p_b) &&\text{for at least one }j.
\end{aligned}
\label{eq:dominance}
\end{equation}
Writing $e_a\prec e_b$ for the corresponding records, the nondominated set is
\begin{equation}
F=\{e\in M^+:\nexists e'\in M^+\text{ such that }e'\prec e\}.
\label{eq:psa-front}
\end{equation}
This set represents trade-offs among the retained recent evaluations, rather than the complete search history. PSA combines high-scoring nondominated records with a small number of recent dominated records, pairing their metrics with module summaries. Appendix~\ref{sec:archive-details} specifies memory updates and retrieval settings.

The resulting summary $b_n$ is saved with the candidate and supplied when it later becomes a parent. PSA provides strategy information for revision; parent and final-program selection continue to follow the database rules and scalar score. It therefore supports reuse without assuming recovery of a globally optimal Pareto frontier. Appendix~\ref{sec:case-psa} shows a recorded strategy summary and the LLM response proposing its reuse.

\section{Experiments}
\label{sec:experiment}
Section~\ref{sec:setup} describes the experimental setup. Section~\ref{sec:control-results} compares the selected evolved programs with reference controllers, and Section~\ref{sec:transfer} examines their transfer across road networks and traffic demands. Section~\ref{sec:search-results} evaluates evolution performance and component ablations. Finally, Section~\ref{sec:objective-results} analyzes how objective weights affect program performance and the observed search distributions.

\subsection{Experimental Setup}
\label{sec:setup}
\subsubsection{Traffic Networks and Simulation Environment}
\begin{figure}[!htbp]
\centering
\includegraphics[width=\linewidth]{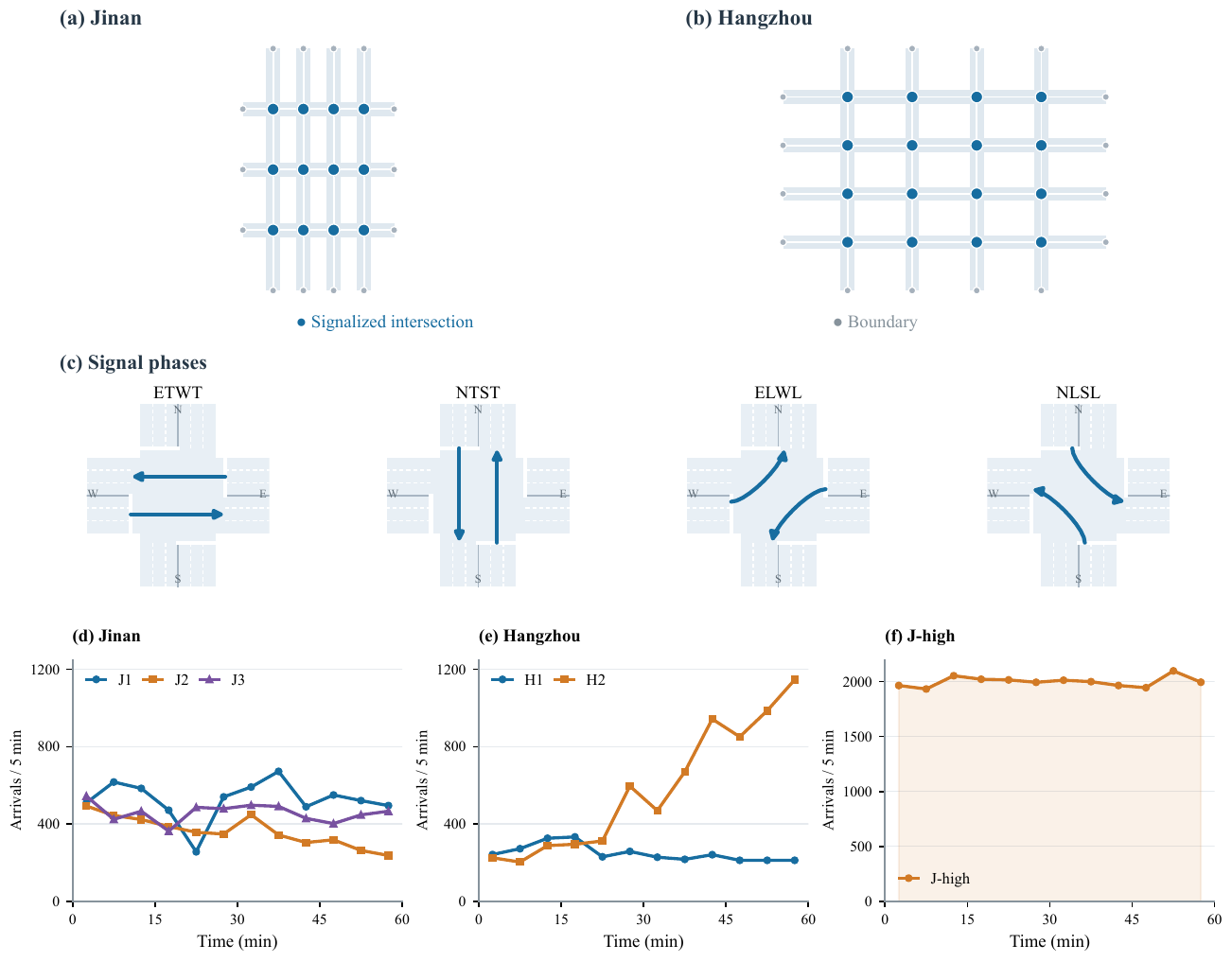}
\caption{Traffic networks, selectable phases, and demand profiles. (a,b) Jinan and Hangzhou layouts. (c) Four selectable phases. (d,e) Observed demand on a common scale. (f) J-high on a separate scale. Arrivals use five-minute intervals.}
\label{fig:setup}
\end{figure}

\paragraph{Networks and traffic demand}
We use two real-world road networks and demand profiles from public TSC benchmarks~\citep{wei2019survey,tscbenchmark}, with roadnet and flow files distributed by LLMLight~\citep{lai2025llmlight}. Jinan's Dongfeng network contains 12 signalized intersections, and Hangzhou's Gudang network contains 16. Each approach has three incoming lanes for left-turn, through, and right-turn traffic. Figure~\ref{fig:setup} shows the layouts, phases, and demand profiles; Table~\ref{tab:scenarios} defines all scenario identifiers.

The main search uses J1, and the principal control comparison covers the five observed-demand scenarios J1--J3 and H1--H2. An auxiliary search on H1 examines transfer in the reverse direction. J-high, H-high, and J-24h provide supplementary tests under synthetic high and day-long demand, reported in Appendix~\ref{sec:frozen-appendix}.

\begin{table}[!htbp]
\centering\small
\caption{Traffic demand statistics for the two road networks.}
\label{tab:scenarios}
\setlength{\tabcolsep}{3pt}
\renewcommand{\arraystretch}{1.13}
\begin{tabular}{@{}llrrrrrrr@{}}
\toprule
\multirow{2}{*}{Demand scenario} & \multirow{2}{*}{ID} & \multirow{2}{*}{\shortstack{Signalized\\intersections}} & \multirow{2}{*}{Hours} & \multirow{2}{*}{Vehicles} & \multicolumn{4}{c}{Arrivals (vehicles / 5 min)} \\
\cmidrule(l){6-9}
 & & & & & Mean & SD & Min & Max \\
\midrule
\multicolumn{9}{@{}l}{\textit{Core demand profiles}} \\
Jinan 1 & J1 & \multirow{4}{*}{12} & 1 & 6,295 & 524.58 & 98.53 & 256 & 672 \\
Jinan 2 & J2 &  & 1 & 4,365 & 363.75 & 74.66 & 237 & 493 \\
Jinan 3 & J3 &  & 1 & 5,494 & 457.83 & 46.22 & 363 & 544 \\
Jinan Extreme$^{\dagger}$ & J-high &  & 1 & 24,000 & 2000.00 & 44.21 & 1,933 & 2,098 \\
\addlinespace[3pt]
Hangzhou 1 & H1 & \multirow{2}{*}{16} & 1 & 2,983 & 248.58 & 40.45 & 212 & 333 \\
Hangzhou 2 & H2 &  & 1 & 6,984 & 582.00 & 318.52 & 203 & 1,146 \\
\midrule
\multicolumn{9}{@{}l}{\textit{Supplementary demand profiles}} \\
Hangzhou Extreme$^{\dagger}$ & H-high & 16 & 1 & 24,000 & 2000.00 & 40.79 & 1,934 & 2,074 \\
Jinan 24-hour$^{\dagger}$ & J-24h & 12 & 24 & 58,200 & 202.08 & 160.88 & 3 & 570 \\
\bottomrule
\end{tabular}

\par\smallskip\begin{minipage}{\linewidth}\scriptsize $\dagger$: synthetic demand. Statistics are computed from scheduled departures in the actual flow files, using consecutive, non-overlapping five-minute bins over the full horizon. SD is the population standard deviation across these bins; vehicle totals are scheduled departures, not completed trips.\end{minipage}
\end{table}

\paragraph{Simulation and signal control}
Experiments use CityFlow~\citep{zhang2019cityflow} through the LLMLight environment~\citep{lai2025llmlight}, with 1-s simulation steps. Controllers choose among east--west through (ETWT), north--south through (NTST), east--west left-turn (ELWL), and north--south left-turn (NLSL) phases every 30~s. Changing phase incurs a 5-s yellow transition within this interval; retaining the current phase incurs none. Right turns remain available. The environment enforces this timing independently of program-level priority adjustments.

\paragraph{Evaluation metrics}
We report travel time (TT), queue length (QL), and waiting time (WT), all minimized. Their definitions follow Section~\ref{sec:problem}.

\subsubsection{Compared Methods}
We compare with 12 reference controllers and two LLM-guided program evolution frameworks, and evaluate four component ablations.

\noindent\textbf{Conventional controllers.} Three conventional policies provide reference performance.
\begin{itemize}\setlength{\itemsep}{2pt}\setlength{\parskip}{0pt}
\item \textbf{Random.} Selects an admissible phase at random.
\item \textbf{FixedTime.} Follows a predetermined phase schedule with fixed service durations.
\item \textbf{Maxpressure.} Selects the phase with the largest pressure, using upstream and downstream queue differences to prioritize service~\citep{varaiya2013max}.
\end{itemize}

\noindent\textbf{RL-based controllers.} Six methods cover pressure-based learning, attention-based phase selection, and coordination between intersections.
\begin{itemize}\setlength{\itemsep}{2pt}\setlength{\parskip}{0pt}
\item \textbf{MPLight.} Combines pressure-based state and reward information with a phase-competition network and shared policy parameters~\citep{chen2020toward}.
\item \textbf{AttendLight.} Aggregates lane information using attention and selects an available phase~\citep{oroojlooy2020attendlight}.
\item \textbf{PressLight.} Incorporates intersection pressure into RL to support network traffic control~\citep{wei2019presslight}.
\item \textbf{CoLight.} Uses graph attention to exchange information among neighboring intersections~\citep{wei2019colight}.
\item \textbf{Efficient-CoLight.} Augments CoLight with an efficient-pressure representation of traffic movements~\citep{wu2021efficient}.
\item \textbf{Advanced-CoLight.} Represents both queuing pressure and approaching traffic through advanced traffic-state features~\citep{zhang2022expression}.
\end{itemize}

\noindent\textbf{LLM-based controllers.} Three methods use LLM inference to produce signal decisions. GPT-4 and LightGPT (Llama2-13B) are the strongest generalist and traffic-specialized variants, respectively, across all five scenarios and three metrics in the reported LLMLight benchmark~\citep{lai2025llmlight}.
\begin{itemize}\setlength{\itemsep}{2pt}\setlength{\parskip}{0pt}
\item \textbf{LLMLight (GPT-4).} Uses GPT-4 to select phases from descriptions of observed traffic~\citep{lai2025llmlight}.
\item \textbf{LightGPT (Llama2-13B).} Uses a traffic-specialized Llama2-13B model developed through imitation fine-tuning and critic-guided policy refinement~\citep{lai2025llmlight}.
\item \textbf{Traffic-R1.} Applies two-stage reinforcement learning post-training to Qwen2.5-3B, using expert-guided offline learning followed by interaction with simulated traffic environments~\citep{zou2025traffic}. At deployment, the resulting LLM processes traffic observations and generates signal-phase decisions.
\end{itemize}

\noindent\textbf{LLM-guided program evolution frameworks.} Both frameworks use the same 3M control interface. We compare their selected programs and their search performance, referring to the adapted implementations as OpenEvolve and ShinkaEvolve.
\begin{itemize}\setlength{\itemsep}{2pt}\setlength{\parskip}{0pt}
\item \textbf{OpenEvolve.} Adapts an open-source framework for LLM-guided program evolution with execution-based feedback~\citep{openevolve}.
\item \textbf{ShinkaEvolve.} Combines LLM-based program revisions with evolutionary search and adaptive proposal mechanisms~\citep{lange2025shinka}.
\end{itemize}

\paragraph{EvoSignal configurations.}
\label{sec:evosignal-variants}
We evaluate three configurations of EvoSignal, distinguished by their objective weights and source scenarios:
\begin{itemize}\setlength{\itemsep}{2pt}\setlength{\parskip}{0pt}
\item \textbf{EvoSignal.} Evolves programs on J1 using the combined-score objective in Eq.~\eqref{eq:fitness}, with $(w_T,w_Q,w_W)=(0.4,0.3,0.3)$ for travel time (TT), queue length (QL), and waiting time (WT), respectively.
\item \textbf{EvoSignal (TT--QL).} Uses the same objective with $(w_T,w_Q,w_W)=(0.48,0.48,0.04)$, increasing the emphasis on TT and QL and reducing the weight assigned to WT. Evolution is also conducted on J1.
\item \textbf{EvoSignal (H1, TT--QL).} Uses the same TT--QL weights but evolves programs on H1 to examine transfer in the reverse direction.
\end{itemize}
Each configuration supplies a fixed program for the five-scenario comparison. We compare traffic metrics rather than scores computed under different objectives.

\noindent\textbf{Ablation variants.} We evaluate four variants of EvoSignal, each removing one component from the full framework.
\begin{itemize}\setlength{\itemsep}{2pt}\setlength{\parskip}{0pt}
\item \textbf{Without 3M.} Represents the controller as a single editable code block and removes module-specific revision guidance.
\item \textbf{Without Multi-init.} Initializes evolution with only the pressure--demand program in Table~\ref{tab:initial-strategies}, instead of the full set of 11 initial programs.
\item \textbf{Without SPA.} Removes traffic diagnostics and revision suggestions from the feedback supplied to the LLM, while retaining numerical evaluation results.
\item \textbf{Without PSA.} Removes the Pareto strategy memory and its summaries from revision prompts, while retaining the underlying program database.
\end{itemize}

\subsubsection{Evolution Settings and Evaluation Protocol}
\label{sec:protocol}
\paragraph{Control results and program selection}

The control tables use 11 reported LLMLight-benchmark controllers~\citep{lai2025llmlight} and Traffic-R1~\citep{zou2025traffic}, whose results average three evaluations per scenario. For each J1 configuration, we select the returned program with the highest combined score on J1 across independent runs; an H1 program provides the reverse-transfer comparison. Each is evaluated unchanged on all five scenarios, without target-specific tuning or reselection, and ranked independently against the 12 references. Rank is one plus the number of references with a strictly lower metric value.

\paragraph{Initial control programs}
Following Section~\ref{sec:multi-init}, we construct 11 seeds with LLM assistance from pressure-based control, feedback control, service fairness, congestion management, and coordination principles. Each uses the 3M interfaces and is checked for successful execution before evolution. Appendix~\ref{sec:initial-program-details} lists their design principles and control rules; the pressure--demand seed alone initializes the without-Multi-init ablation.

\paragraph{Evolution budget and repeated runs}
Full EvoSignal starts from 11 programs and runs 200 evolution iterations after initialization. Following AlphaEvolve's use of fast and more capable models~\citep{novikov2025alphaevolve}, EvoSignal and its ablations use DeepSeek-V4-Flash and DeepSeek-V4-Pro~\citep{deepseek2026v4}, sampled with probabilities 70\% and 30\%, respectively, both at temperature 0.7. The primary framework and ablation comparisons use three independent runs per configuration; J1 TT--QL uses two and H1 TT--QL uses three. Appendix~\ref{sec:implementation-settings} provides the remaining settings.

Framework comparisons measure progress toward higher combined scores, final score quality, and variation across runs. The framework comparison uses progress relative to each framework's configured iteration or generation budget; the ablation comparison uses evolution iterations. Neither axis measures elapsed time. Fixed control programs are evaluated separately by TT, QL, and WT across traffic scenarios.

\subsection{Traffic Control Performance of Selected Evolved Programs}
\label{sec:control-results}
Tables~\ref{tab:control_jinan} and~\ref{tab:control_hangzhou} report TT, QL, and WT for the selected fixed control programs on the Jinan and Hangzhou networks. For EvoSignal, these are the programs selected by their source-scenario combined scores after 200 evolution iterations, following Section~\ref{sec:protocol}. Each program is ranked independently against the same 12 reference controllers. For each scenario and metric, the best reference value is the lowest value achieved by any of the 12 reference controllers.

\begin{table}[!htbp]
\centering\footnotesize
\caption{Traffic control performance on Jinan. Lower metric values and ranks are better.}
\label{tab:control_jinan}
\setlength{\tabcolsep}{3.5pt}
\begin{tabular}{lrrrrrrrrr}
\toprule
Method & \multicolumn{3}{c}{J1} & \multicolumn{3}{c}{J2} & \multicolumn{3}{c}{J3} \\
 & TT (s) & QL (veh) & WT (s) & TT (s) & QL (veh) & WT (s) & TT (s) & QL (veh) & WT (s) \\
\midrule
\rowcolor[gray]{0.92}\multicolumn{10}{l}{\textit{Conventional controllers}} \\
Random & 597.62 & 687.35 & 99.46 & 555.23 & 428.38 & 100.40 & 552.74 & 529.63 & 99.33 \\
FixedTime & 481.79 & 491.03 & 70.99 & 441.19 & 294.14 & 66.72 & 450.11 & 394.34 & 69.19 \\
Maxpressure & 281.58 & 170.71 & 44.53 & 273.20 & 106.58 & \textbf{38.25} & 265.75 & 133.90 & 40.20 \\
\rowcolor[gray]{0.92}\multicolumn{10}{l}{\textit{RL-based controllers}} \\
MPLight & 307.82 & 215.93 & 97.88 & 304.51 & 142.25 & 90.91 & 291.79 & 171.70 & 89.93 \\
AttendLight & 291.29 & 186.25 & 61.73 & 280.94 & 115.52 & 52.46 & 273.02 & 144.05 & 55.93 \\
PressLight & 291.57 & 185.46 & 50.53 & 281.46 & 115.99 & 47.27 & 275.85 & 148.18 & 54.81 \\
CoLight & 279.60 & 168.53 & 58.87 & 274.77 & 108.28 & 54.14 & 266.39 & 135.08 & 53.33 \\
Efficient-CoLight & 277.11 & 163.60 & 43.41 & 269.24 & 102.98 & 39.74 & 262.25 & 129.72 & \textbf{39.99} \\
Advanced-CoLight & 274.67 & 160.85 & 49.30 & 268.25 & 102.12 & 41.11 & 260.66 & 127.83 & 43.54 \\
\rowcolor[gray]{0.92}\multicolumn{10}{l}{\textit{LLM-based controllers}} \\
LLMLight (GPT-4) & 275.26 & 160.93 & 46.61 & 271.34 & 105.22 & 47.55 & 264.70 & 132.53 & 46.16 \\
LightGPT (Llama2-13B) & \textbf{274.03} & \textbf{159.39} & \textbf{43.24} & \textbf{266.94} & \textbf{100.46} & 40.34 & \textbf{260.17} & \textbf{127.08} & 41.00 \\
Traffic-R1 & 277.92 & 165.13 & 44.21 & 268.90 & 102.32 & 41.00 & 262.73 & 130.21 & 41.89 \\
\rowcolor[gray]{0.92}\multicolumn{10}{l}{\textit{Evolved control programs}} \\
\rowcolor[gray]{0.97}\textbf{OpenEvolve} & 278.20 & 163.38 & \textbf{34.36} & 268.30 & 101.28 & \textbf{32.03} & 261.86 & 128.40 & \textbf{33.18} \\
\quad\textit{Rank vs.\ references} & 6 & 4 & 1 & 3 & 2 & 1 & 3 & 3 & 1 \\
\rowcolor[gray]{0.97}\textbf{ShinkaEvolve} & 282.38 & 168.91 & \textbf{32.64} & 267.16 & \textbf{100.39} & \textbf{29.76} & 262.49 & 129.41 & \textbf{31.91} \\
\quad\textit{Rank vs.\ references} & 8 & 7 & 1 & 2 & 1 & 1 & 4 & 3 & 1 \\
\rowcolor[rgb]{0.91,0.95,0.97}\textbf{EvoSignal} & 277.74 & 163.29 & \textbf{34.31} & 268.52 & 102.30 & \textbf{31.71} & 262.20 & 129.09 & \textbf{33.28} \\
\quad\textit{Rank vs.\ references} & 5 & 4 & 1 & 3 & 3 & 1 & 3 & 3 & 1 \\
\rowcolor[rgb]{0.91,0.95,0.97}\textbf{EvoSignal (TT--QL)} & \textbf{271.85} & \textbf{155.53} & \textbf{41.98} & \textbf{266.94} & \textbf{100.08} & 38.40 & 260.58 & 127.79 & \textbf{39.72} \\
\quad\textit{Rank vs.\ references} & 1 & 1 & 1 & 1 & 1 & 2 & 2 & 2 & 1 \\
\rowcolor[rgb]{0.91,0.95,0.97}\textbf{EvoSignal (H1, TT--QL)} & 274.51 & 159.81 & \textbf{40.32} & 268.34 & 102.08 & \textbf{36.61} & \textbf{259.95} & \textbf{126.13} & \textbf{36.88} \\
\quad\textit{Rank vs.\ references} & 2 & 2 & 1 & 3 & 2 & 1 & 1 & 1 & 1 \\
\bottomrule
\end{tabular}

\par\smallskip\begin{minipage}{\linewidth}\scriptsize References comprise 11 reported LLMLight-benchmark controllers~\citep{lai2025llmlight} and Traffic-R1 (mean of three evaluations). Each evolved program is ranked independently against these 12 references (1--13); other evolved programs are excluded. Bold marks the best reference values and evolved values that match or improve on them. Ranks use the displayed two-decimal metrics. Blue rows denote EvoSignal. --: unavailable. EvoSignal (H1, TT--QL) was searched on H1; the other evolved programs were searched on J1. Objective variants are defined in Section~\ref{sec:evosignal-variants}.\end{minipage}
\end{table}

\begin{table}[!htbp]
\centering\footnotesize
\caption{Traffic control performance on Hangzhou. Lower metric values and ranks are better.}
\label{tab:control_hangzhou}
\setlength{\tabcolsep}{3.5pt}
\begin{tabular}{lrrrrrr}
\toprule
Method & \multicolumn{3}{c}{H1} & \multicolumn{3}{c}{H2} \\
 & TT (s) & QL (veh) & WT (s) & TT (s) & QL (veh) & WT (s) \\
\midrule
\rowcolor[gray]{0.92}\multicolumn{7}{l}{\textit{Conventional controllers}} \\
Random & 621.14 & 295.81 & 96.06 & 504.28 & 432.92 & 92.61 \\
FixedTime & 616.02 & 301.33 & 73.99 & 486.72 & 425.15 & 72.80 \\
Maxpressure & 325.33 & 68.99 & 49.60 & 347.74 & 215.53 & 70.58 \\
\rowcolor[gray]{0.92}\multicolumn{7}{l}{\textit{RL-based controllers}} \\
MPLight & 345.60 & 84.70 & 81.97 & 358.56 & 237.17 & 100.16 \\
AttendLight & 322.94 & 66.96 & 55.19 & 358.81 & 239.05 & 72.88 \\
PressLight & 364.13 & 98.67 & 90.33 & 417.01 & 349.25 & 150.46 \\
CoLight & 322.85 & 66.94 & 61.82 & 342.90 & 212.09 & 99.74 \\
Efficient-CoLight & 311.96 & 58.20 & \textbf{36.83} & 333.27 & 189.65 & \textbf{61.70} \\
Advanced-CoLight & \textbf{304.47} & \textbf{52.94} & 41.75 & \textbf{329.16} & 186.34 & 76.59 \\
\rowcolor[gray]{0.92}\multicolumn{7}{l}{\textit{LLM-based controllers}} \\
LLMLight (GPT-4) & 318.71 & 62.84 & 58.09 & 335.81 & 193.32 & 66.02 \\
LightGPT (Llama2-13B) & 310.78 & 56.93 & 38.64 & 330.71 & 189.09 & 64.16 \\
Traffic-R1 & 310.66 & 57.14 & 42.65 & 330.30 & \textbf{184.79} & 62.46 \\
\rowcolor[gray]{0.92}\multicolumn{7}{l}{\textit{Evolved control programs}} \\
\rowcolor[gray]{0.97}\textbf{OpenEvolve} & 310.55 & 57.66 & \textbf{29.39} & \textbf{328.98} & \textbf{178.87} & \textbf{36.22} \\
\quad\textit{Rank vs.\ references} & 2 & 4 & 1 & 1 & 1 & 1 \\
\rowcolor[gray]{0.97}\textbf{ShinkaEvolve} & 315.67 & 60.90 & \textbf{27.93} & 329.30 & \textbf{176.78} & \textbf{30.95} \\
\quad\textit{Rank vs.\ references} & 5 & 5 & 1 & 2 & 1 & 1 \\
\rowcolor[rgb]{0.91,0.95,0.97}\textbf{EvoSignal} & 315.43 & 61.24 & \textbf{29.59} & \textbf{327.82} & \textbf{174.61} & \textbf{31.32} \\
\quad\textit{Rank vs.\ references} & 5 & 5 & 1 & 1 & 1 & 1 \\
\rowcolor[rgb]{0.91,0.95,0.97}\textbf{EvoSignal (TT--QL)} & 307.92 & 56.64 & 38.53 & 329.81 & 185.67 & \textbf{61.53} \\
\quad\textit{Rank vs.\ references} & 2 & 2 & 2 & 2 & 2 & 1 \\
\rowcolor[rgb]{0.91,0.95,0.97}\textbf{EvoSignal (H1, TT--QL)} & 307.33 & 55.34 & \textbf{31.15} & \textbf{328.68} & \textbf{180.92} & \textbf{54.70} \\
\quad\textit{Rank vs.\ references} & 2 & 2 & 1 & 1 & 1 & 1 \\
\bottomrule
\end{tabular}

\par\smallskip\begin{minipage}{\linewidth}\scriptsize Ranking, boldface, and color conventions follow Table~\ref{tab:control_jinan}. Each evolved program is compared independently with the same 12 references. EvoSignal (H1, TT--QL) was searched on H1; the others were searched on J1.\end{minipage}
\end{table}

The selected programs from all three evolution frameworks achieve lower WT than the best reference value in all five scenarios. The selected default EvoSignal program reduces WT by 16.78--49.24\%, including on demands and a road network not used for search. TT and QL vary more across evolved programs.

On H2, the selected default EvoSignal program outperforms all 12 reference controllers on all three metrics. The EvoSignal (TT--QL) program improves on every reference controller on all three metrics in J1 and ranks first against these references on TT and QL in J2, with TT tied at the reported precision. These results show complementary performance profiles under different design objectives, examined in Section~\ref{sec:objective-results}. Section~\ref{sec:transfer} examines transfer in both directions for EvoSignal (TT--QL) and EvoSignal (H1, TT--QL).

\begin{figure}[!tp]
\centering
\includegraphics[width=\linewidth]{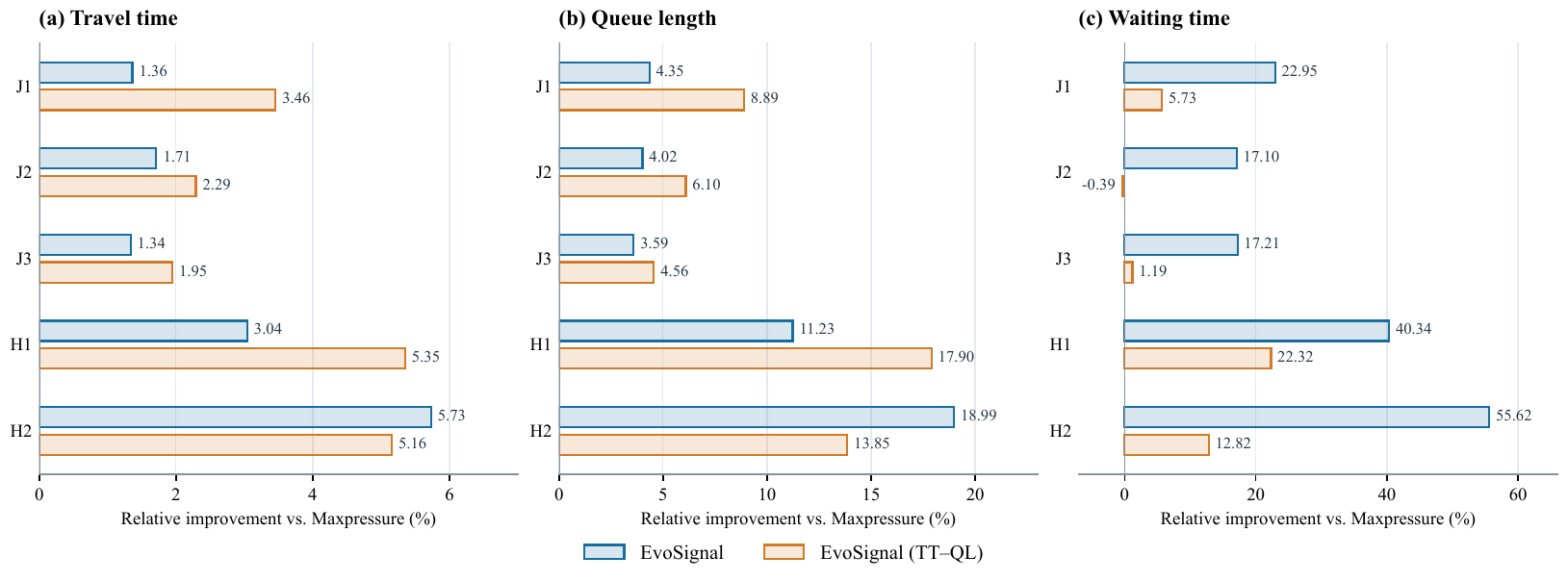}
\caption{Relative improvement over Maxpressure for two EvoSignal programs evolved on J1 and reused unchanged across five scenarios. Positive values indicate improvement.}
\label{fig:objective-tradeoffs}
\end{figure}

\paragraph{Performance relative to Maxpressure}
Figure~\ref{fig:objective-tradeoffs} highlights performance gains over Maxpressure. Like the evolved programs, Maxpressure uses explicit control rules and lightweight online computation. This comparison therefore assesses improvements within the same white-box, computationally lightweight form of control. Relative improvement is $100(m_{\mathrm{MP}}-m_{\mathrm{Evo}})/m_{\mathrm{MP}}$ for metric $m$. The default EvoSignal program improves all three metrics in every scenario, with WT reductions of 17.10--55.62\%. The TT--QL program improves TT and QL further on J1--J3 and H1, but its WT on J2 is 0.39\% higher than Maxpressure. The main tables retain comparisons with all 12 reference controllers.

\subsection{Transfer across Road Networks and Traffic Demands}
\label{sec:transfer}
\paragraph{Transfer in both directions}
All five scenarios share the observation structure and four-phase control interface described in Section~\ref{sec:setup}. Tables~\ref{tab:control_jinan} and~\ref{tab:control_hangzhou} compare EvoSignal (TT--QL), evolved on J1, with EvoSignal (H1, TT--QL), evolved on H1 under the same objective weights. EvoSignal (TT--QL) is reused on J2 and J3, and on the different road network in H1 and H2. Conversely, EvoSignal (H1, TT--QL) is reused on H2 and on all three Jinan scenarios. Neither program undergoes further search or target-specific tuning. Both programs rank within the top three against the 12 reference controllers on every metric across the five scenarios. EvoSignal (H1, TT--QL) outperforms all 12 references on TT, QL, and WT in both J3 and H2. These results support reuse of the evolved control logic across both road networks and traffic demands.

\paragraph{Influence of the search scenario}
On J1, EvoSignal (TT--QL) reduces TT and QL by 0.97\% and 2.68\% relative to EvoSignal (H1, TT--QL), while WT is 4.12\% higher. On H1, EvoSignal (H1, TT--QL) improves TT, QL, and WT by 0.19\%, 2.30\%, and 19.14\%, respectively, relative to EvoSignal (TT--QL). Source-scenario advantages are not uniform: EvoSignal (H1, TT--QL) also improves all three metrics on J3. Thus, strong transfer can coexist with scenario-dependent performance, and differences between evolved rules cannot be attributed to the search scenario alone.

Supplementary evaluations under high and day-long demand show that these performance advantages do not extend uniformly to heavier traffic. Appendix~\ref{sec:frozen-appendix} reports the complete fixed-program group results and their between-program variation.

\subsection{Evolution Performance and Component Ablations}
\label{sec:search-results}
\begin{figure}[!htbp]
\centering
\includegraphics[width=\linewidth]{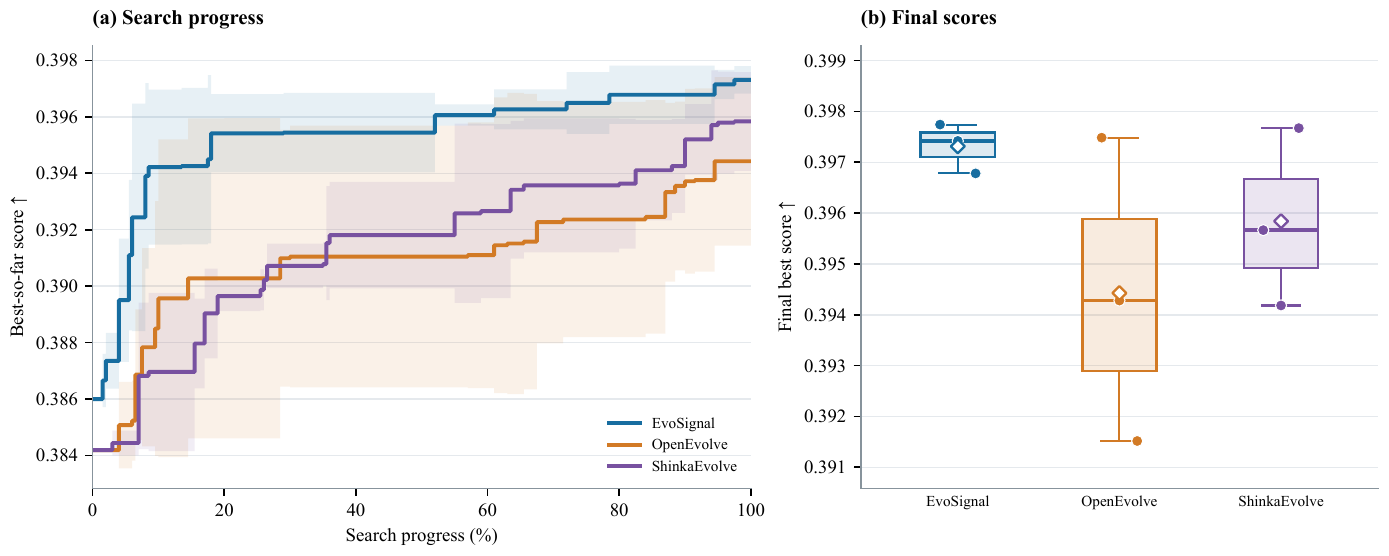}
\caption{Evolution-framework comparison on J1 under the default objective. (a) Mean best-so-far score with SD bands over three runs. (b) Distribution of final best scores.}
\label{fig:framework-search}
\end{figure}

\paragraph{Comparison with evolution frameworks}
Figure~\ref{fig:framework-search} compares the three evolution frameworks on J1; Appendix~\ref{sec:sensitivity} provides their exact final-score statistics. Curves show the mean best-so-far combined score during evolution over three independent runs, with bands indicating sample standard deviations (SDs). EvoSignal achieves a mean final best score of $0.397314\pm0.000489$, compared with $0.394427\pm0.002986$ for OpenEvolve and $0.395840\pm0.001751$ for ShinkaEvolve. It has the highest mean score and the lowest variation across these runs.

\begin{figure}[!htbp]
\centering
\includegraphics[width=\linewidth]{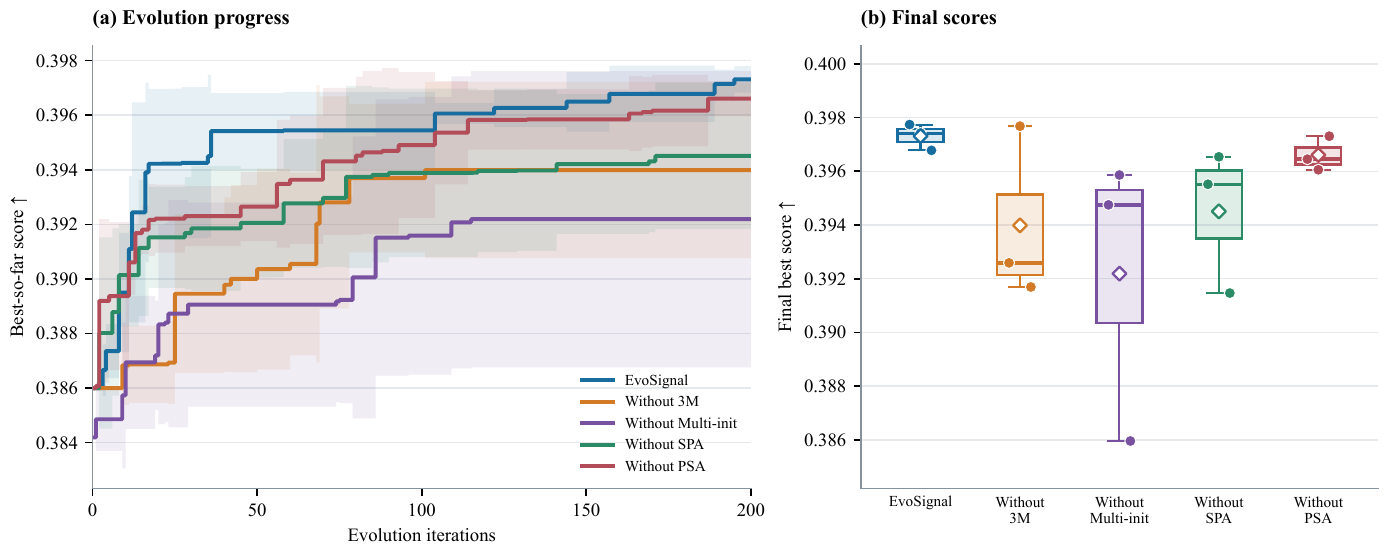}
\caption{Component ablations on J1 under the default objective: (a) best-so-far score over 200 evolution iterations; (b) final best scores. Each setting uses three runs, with the same plotting conventions as Figure~\ref{fig:framework-search}.}
\label{fig:module-ablation}
\end{figure}

\paragraph{Component ablations}
\label{sec:ablation-results}
Figure~\ref{fig:module-ablation} compares the four ablations. Full EvoSignal has the highest mean final score and lowest final SD among these settings; Table~\ref{tab:search} in Appendix~\ref{sec:sensitivity} reports the values.

\paragraph{Without 3M.}
The monolithic controller reaches a similar best single-run score to full EvoSignal but has a lower mean score and substantially greater variation across runs. This supports more consistent search with 3M and its revision guidance. Explicit interfaces also support inspection and targeted revision around available sensing and communication resources.

\paragraph{Without Multi-init.}
Removing Multi-init yields the lowest mean final score, $0.392190\pm0.005428$. Full EvoSignal's three final programs descend from different seeds---fairness-aware, efficient-pressure, and corridor-inspired control---showing use of multiple initial ideas. Initial scores and valid-record differences are reported in Appendix~\ref{sec:sensitivity}.

\paragraph{Without SPA.}
Without SPA, the mean final score falls to 0.394508 and SD rises to 0.002684, consistent with the intended role of traffic diagnostics. Appendix~\ref{sec:revision-example} illustrates how diagnostics and archived strategies enter a recorded revision.

\paragraph{Without PSA.}
Removing PSA has a smaller effect, yielding $0.396607\pm0.000639$.

\subsection{Influence of Objective Weights}
\label{sec:objective-results}
\paragraph{Objective weights express operational priorities}
The two J1 programs in Figure~\ref{fig:objective-tradeoffs} also illustrate the effect of objective weights. On J1, the TT--QL program reduces TT from 277.74 to 271.85~s and QL from 163.29 to 155.53 vehicles relative to the default program, while WT rises from 34.31 to 41.98~s. The weights and metric scales in Eq.~\eqref{eq:fitness} determine how these changes contribute to the combined score. Increasing the TT and QL weights reverses the ranking of these two programs; Appendix~\ref{sec:pareto-full} gives the score decomposition. Scores computed under different objectives are not directly comparable.

\begin{figure}[!htbp]
\centering
\includegraphics[width=\linewidth]{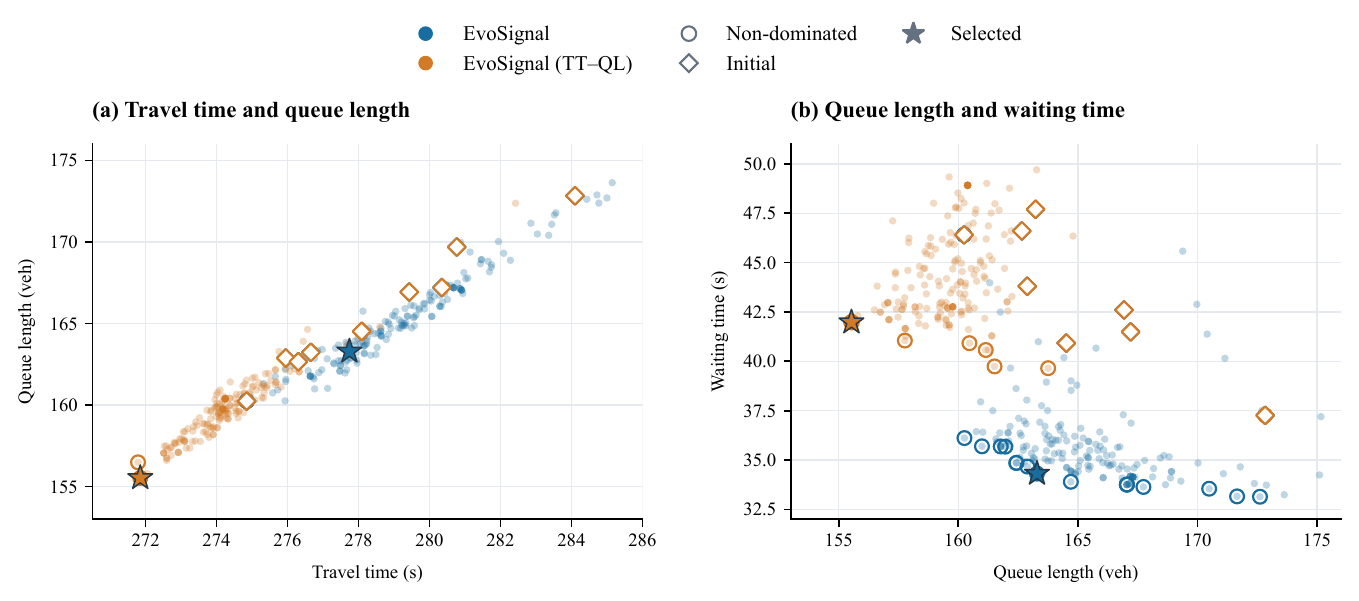}
\caption{Objective-space distributions under the default and TT--QL objectives on J1: (a) travel time versus queue length; (b) queue length versus waiting time. Additional views and plotting details are provided in Appendix~\ref{sec:pareto-full}.}
\label{fig:objective-pareto}
\end{figure}

\paragraph{Weights also change the observed search distribution}
Figure~\ref{fig:objective-pareto} shows the searches producing these programs. The TT--QL search concentrates on lower TT and QL, while the default search reaches lower WT; their selected programs occupy different parts of this trade-off. These observations describe the two recorded searches, rather than an optimal weight setting. Appendix~\ref{sec:pareto-full} reports distribution statistics and additional views.

The TT--QL program also has lower TT and QL but higher WT on J2, J3, and H1. On H2, the default program performs better on all three metrics. Objective weights guide the design of a new program rather than changing a fixed program's decisions; the resulting trade-offs must therefore be assessed under the intended traffic conditions.

\section{Program Analysis and Discussion}
\label{sec:discussion}
\subsection{Control Rules of the Selected Programs}
\label{sec:selected-rules}
The two programs evolved on J1 and evaluated in Tables~\ref{tab:control_jinan} and~\ref{tab:control_hangzhou} illustrate the executable rules produced by EvoSignal. Both follow the same decision process: M1 extracts phase-level traffic features, M2 computes a priority for each candidate phase, and M3 supplies an optional network-based adjustment before the controller selects the phase with the highest priority. Table~\ref{tab:selected-rules} summarizes the rules in their three modules.

Both programs use pressure, waiting vehicles, approaching vehicles, and recent phase-service history. Their main differences lie in how M2 combines this information. Default EvoSignal increases compensation for phases not recently requested as their waiting demand grows, while reducing the pressure contribution when downstream traffic is heavy. EvoSignal (TT--QL) adjusts the weight of waiting vehicles according to intersection congestion, limits service-age compensation, and reduces the combined priority when downstream occupancy is high. It also adds a preference for continuing the current phase when approaching demand remains relatively high.

The explicit program representation makes the implemented decision rules available for inspection. They provide a concrete interpretation of the complementary performance profiles in Section~\ref{sec:objective-results}; the aggregate results do not isolate the contribution of each rule. Appendix~\ref{sec:selected-programs} details the feature definitions, priority calculations, conditional rules, coefficients, and service-history updates. Appendix~\ref{sec:editable-program} shows the corresponding editable code structure.

\begin{table}[!htbp]
\centering\footnotesize
\caption{Control rules in the three modules of the two selected J1 programs. Details are provided in Appendix~\ref{sec:selected-programs}.}
\label{tab:selected-rules}
\begin{tabular}{@{}p{0.16\linewidth}p{0.39\linewidth}p{0.39\linewidth}@{}}
\toprule
Component & EvoSignal & EvoSignal (TT--QL) \\
\midrule
M1: features & Pressure, waiting and approaching traffic, upstream occupancy, downstream load, and time since a phase was requested. & Pressure, waiting and approaching traffic, maximum downstream occupancy, and time since a phase was observed active. \\
M2: service & Service-age compensation grows with both age and waiting demand. & Service-age compensation is capped; the waiting weight adapts to congestion. \\
M2: congestion & Downstream load attenuates the pressure term; upstream occupancy amplifies demand. & High downstream occupancy attenuates the combined phase score. \\
M2: persistence & Applies different discounts to current and competing phase priorities at regular decision points. & Adds a continuation bonus when approaching demand remains relatively high. \\
M3: network adjustment & Favors the through phase with greater network directional pressure; adds a common offset based on neighboring intersection loads. & No additional adjustment. \\
\bottomrule
\end{tabular}
\end{table}

\subsection{How Existing Strategies Are Revised}
\label{sec:program-analysis}
The default program provides an example of how evolution extends an initial control idea. Its fairness-aware seed, summarized in Table~\ref{tab:initial-strategies}, increases priority for phases not recently selected and adjusts the preference for retaining the current phase. Evolution retains this service-history idea while incorporating pressure, waiting and approaching demand, and downstream traffic conditions. Later revisions change the demand weights and strengthen compensation for phases that have both a long service age and many waiting vehicles. Thus, the final rule combines service fairness with current traffic demand and downstream conditions. Appendix~\ref{sec:default-revisions} traces the code changes and their measured outcomes.

Appendix~\ref{sec:revision-example} provides a separate recorded example showing how such edits are produced: traffic diagnostics and archived strategies enter the prompt, and the LLM revises feature extraction and phase scoring through SEARCH/REPLACE edits. The appendix links the input, response, saved code, and evaluation results. Together, these records illustrate changes to the structure and use of traffic features, beyond adjusting coefficients in a fixed rule; they do not establish that each revision improves every metric or achieves its stated intent.

\subsection{Implications for Traffic Control Design}
The four components address different needs in automated traffic control design. 3M separates the processing of traffic observations from local phase prioritization and optional network-based adjustment. Multi-init supplies several established control ideas as executable starting points. SPA helps identify traffic conditions and service patterns that require attention. PSA makes earlier strategies and their measured trade-offs available for reuse. Together, these components guide the generation and evaluation of control rules whose source code remains available for inspection.

The selected program uses current traffic observations to choose signal phases without online LLM inference. LLM computation is therefore concentrated in the offline design stage, while repeated control decisions execute explicit rules. The experiments demonstrate traffic performance and traceable changes to these rules.

\subsection{Limitations and Future Directions}
\paragraph{Dependence on LLM interpretation and strategy reuse}
SPA computes traffic statistics and rule-based diagnostics, and PSA determines dominance from measured objectives. However, the quality of strategy summaries and the use of diagnostics and archived strategies during revision depend on the LLM. A useful observation may be overlooked or translated into a code change whose behavior differs from the generated explanation. Post-training on traffic control code and verified diagnostic--revision records could improve this capability, but its benefits remain to be evaluated. A complementary direction is to combine language-based strategy reuse with explicit selection and sampling rules that account for objective-space coverage, strategy diversity, and measured performance. Such mechanisms could make the choice of reference programs less dependent on the LLM's interpretation of a limited prompt.

\paragraph{Limited understanding and reuse of evolved control rules}
EvoSignal evaluates complete programs, while a single LLM revision may introduce several interacting changes to features, scoring rules, or conditional logic. A higher program score therefore does not establish which change is useful, under which traffic conditions, or whether its benefit transfers to another program. Future work should remove or replace individual changes while holding the remaining code fixed, examine their interactions across traffic conditions, and test their reuse in other controllers. Inspired by library learning in program synthesis~\citep{wong2021laps}, validated code fragments could be retained with their required inputs, tested conditions, failure cases, and performance trade-offs. Together, these steps could turn useful local improvements into reusable control knowledge and support the construction of better programs in subsequent searches.

\section{Conclusion}
\label{sec:conclusion}
This paper formulated TSC as a modular program design problem and presented EvoSignal to address it through LLM-guided evolution. Executable policies map traffic observations to signal phase decisions through connected modules for feature extraction, local phase prioritization, and optional network-based adjustment. Multiple initial strategies, traffic diagnostics, and reuse of strategies with different performance trade-offs guide the search. Control performance feedback guides code revision during design. The selected program then responds to traffic observations without online LLM inference.

Experiments on real-world road networks and traffic demand show that the selected default EvoSignal program reduces WT by 16.8--49.2\% relative to the lowest WT achieved by the 12 reference controllers in each of the five scenarios. With greater emphasis on TT and QL, the framework produces a program that outperforms all references on all three metrics in J1 and ranks within the top two on every metric across all five scenarios without further revision. EvoSignal also achieves higher mean search scores and lower variation than the evaluated LLM-guided program evolution frameworks. The ablations and program records provide evidence of the roles of the four components. These findings support automated design of explicit traffic control rules, with their suitability assessed against the intended objectives and operating conditions.

\section*{Declaration of Generative AI and AI-assisted technologies in the writing process}
During the preparation of this work, the authors used ChatGPT to improve the clarity and readability of the manuscript. The authors reviewed and edited the content as needed and take full responsibility for the content of the publication.

\par\vspace{8pt}
\begingroup\small
\setlength{\bibsep}{2pt}
\section*{References}
\renewcommand{\bibsection}{} 
\bibliographystyle{unsrtnat}
\bibliography{references}
\endgroup
\clearpage
\appendix
\makeatletter
\let\evosignal@seccntformat\@seccntformat
\def\evosignal@sectionprefix{Appendix~\thesection.\quad}
\renewcommand{\@seccntformat}[1]{%
  \ifcsname evosignal@#1prefix\endcsname
    \csname evosignal@#1prefix\endcsname
  \else
    \evosignal@seccntformat{#1}%
  \fi}
\makeatother
\section{Search Statistics}
\label{sec:sensitivity}
Table~\ref{tab:search} reports the primary three-run comparison used in Figures~\ref{fig:framework-search} and~\ref{fig:module-ablation}.
\begin{table}[!htbp]
\centering\footnotesize
\caption{Final best search scores on J1 under the default objective (Section~\ref{sec:evosignal-variants}).}
\label{tab:search}
\setlength{\tabcolsep}{3.5pt}
\begin{tabular}{lrrr}
\toprule
Configuration & Runs & Mean $\pm$ SD & Best run \\
\midrule
EvoSignal & 3 & 0.397314 $\pm$ 0.000489 & 0.397742 \\
OpenEvolve & 3 & 0.394427 $\pm$ 0.002986 & 0.397484 \\
ShinkaEvolve & 3 & 0.395840 $\pm$ 0.001751 & 0.397672 \\
Without SPA & 3 & 0.394508 $\pm$ 0.002684 & 0.396541 \\
Without PSA & 3 & 0.396607 $\pm$ 0.000639 & 0.397309 \\
Without 3M & 3 & 0.393988 $\pm$ 0.003230 & 0.397682 \\
Without Multi-init & 3 & 0.392190 $\pm$ 0.005428 & 0.395863 \\
\bottomrule
\end{tabular}

\par\smallskip\begin{minipage}{\linewidth}\scriptsize SD is the sample standard deviation across independent search runs. Without 3M uses the archived monolithic controller and its corresponding search guidance.\end{minipage}
\end{table}

\paragraph{Additional component statistics}
The mean best initial score is 0.385995 with Multi-init and 0.384187 with the single seed; subsequent evolution adds 0.011319 and 0.008002 on average, respectively. Single-seed runs also contain fewer valid evolutionary records, which should be considered when interpreting final performance. Averaged over configured search steps, the best-so-far score is 0.395385 with PSA and 0.394389 without it.

\section{Complete Evaluations of Fixed Programs}
\label{sec:frozen-appendix}
Tables~\ref{tab:frozen1} and~\ref{tab:frozen2} report the performance of all fixed program groups across the eight scenarios in Table~\ref{tab:scenarios}. The code is unchanged across scenarios. Means and SDs summarize differences between independently obtained programs within each group. The H1 TT--QL group comprises three independently evolved programs; its group statistics are distinct from the single representative in the main tables.
\paragraph{High and day-long demand}
The three default EvoSignal programs have lower mean WT than the two EvoSignal (TT--QL) programs on J-high, H-high, and J-24h: 197.02 versus 332.66~s, 81.09 versus 312.72~s, and 26.77 versus 29.29~s. Default EvoSignal also has lower TT and QL under both high-flow profiles, while TT--QL has slightly lower TT and QL over the full day. Compared with these two groups, ShinkaEvolve has lower means on all three metrics under both high-flow profiles; on J-24h, default EvoSignal has slightly lower TT and QL, with similar WT. These results show that advantages on the five main scenarios do not extend uniformly to heavier demand. Each fixed program is evaluated once per scenario; SDs describe variation between programs.

\begin{table}[!htbp]
\centering\footnotesize
\caption{Fixed-program results: EvoSignal, EvoSignal (TT--QL).}
\label{tab:frozen1}
\setlength{\tabcolsep}{3.5pt}
\begin{tabular}{lrrrr}
\toprule
Scenario & Programs & TT (s) & QL (veh) & WT (s) \\
\midrule
\multicolumn{5}{l}{\textit{EvoSignal}} \\
J1 & 3 & 275.97 $\pm$ 1.86 & 160.64 $\pm$ 2.50 & 35.64 $\pm$ 1.52 \\
J2 & 3 & 268.21 $\pm$ 0.51 & 101.75 $\pm$ 0.50 & 33.06 $\pm$ 1.43 \\
J3 & 3 & 261.57 $\pm$ 1.00 & 128.56 $\pm$ 0.56 & 34.34 $\pm$ 0.96 \\
H1 & 3 & 316.15 $\pm$ 2.18 & 61.76 $\pm$ 1.32 & 29.88 $\pm$ 0.47 \\
H2 & 3 & 329.69 $\pm$ 1.62 & 177.71 $\pm$ 2.69 & 34.45 $\pm$ 4.92 \\
J-high & 3 & 855.20 $\pm$ 14.10 & 2654.50 $\pm$ 60.38 & 197.02 $\pm$ 66.56 \\
H-high & 3 & 681.27 $\pm$ 26.32 & 1446.06 $\pm$ 88.52 & 81.09 $\pm$ 28.48 \\
J-24h & 3 & 271.31 $\pm$ 1.24 & 56.27 $\pm$ 0.72 & 26.77 $\pm$ 0.35 \\
\multicolumn{5}{l}{\textit{EvoSignal (TT--QL)}} \\
J1 & 2 & 272.27 $\pm$ 0.59 & 156.68 $\pm$ 1.61 & 41.64 $\pm$ 0.48 \\
J2 & 2 & 267.24 $\pm$ 0.42 & 100.50 $\pm$ 0.61 & 37.73 $\pm$ 0.96 \\
J3 & 2 & 260.82 $\pm$ 0.34 & 128.10 $\pm$ 0.44 & 39.73 $\pm$ 0.02 \\
H1 & 2 & 310.94 $\pm$ 4.27 & 58.25 $\pm$ 2.27 & 38.35 $\pm$ 0.25 \\
H2 & 2 & 330.13 $\pm$ 0.45 & 184.25 $\pm$ 2.00 & 56.05 $\pm$ 7.75 \\
J-high & 2 & 902.23 $\pm$ 17.75 & 2870.40 $\pm$ 76.09 & 332.66 $\pm$ 104.67 \\
H-high & 2 & 726.68 $\pm$ 52.65 & 1661.24 $\pm$ 235.47 & 312.72 $\pm$ 275.91 \\
J-24h & 2 & 269.47 $\pm$ 0.44 & 55.18 $\pm$ 0.16 & 29.29 $\pm$ 1.02 \\
\bottomrule
\end{tabular}

\par\smallskip\begin{minipage}{\linewidth}\scriptsize Mean $\pm$ sample SD across different fixed programs. Each program contributes one evaluation to each scenario. EvoSignal (H1, TT--QL) programs were searched on H1; the other groups were searched on J1.\end{minipage}
\end{table}

\begin{table}[!htbp]
\centering\footnotesize
\caption{Fixed-program results: EvoSignal (H1, TT--QL), ShinkaEvolve.}
\label{tab:frozen2}
\setlength{\tabcolsep}{3.5pt}
\begin{tabular}{lrrrr}
\toprule
Scenario & Programs & TT (s) & QL (veh) & WT (s) \\
\midrule
\multicolumn{5}{l}{\textit{EvoSignal (H1, TT--QL)}} \\
J1 & 3 & 276.66 $\pm$ 3.74 & 162.54 $\pm$ 4.26 & 40.44 $\pm$ 1.67 \\
J2 & 3 & 268.40 $\pm$ 0.26 & 102.08 $\pm$ 0.00 & 37.29 $\pm$ 1.36 \\
J3 & 3 & 260.31 $\pm$ 1.06 & 127.36 $\pm$ 1.43 & 38.97 $\pm$ 2.06 \\
H1 & 3 & 307.77 $\pm$ 0.59 & 55.86 $\pm$ 0.59 & 32.34 $\pm$ 2.17 \\
H2 & 3 & 332.12 $\pm$ 3.71 & 186.25 $\pm$ 6.36 & 52.92 $\pm$ 2.34 \\
J-high & 3 & 815.16 $\pm$ 7.94 & 2627.99 $\pm$ 80.90 & 132.63 $\pm$ 39.22 \\
H-high & 3 & 695.38 $\pm$ 11.89 & 1561.01 $\pm$ 15.19 & 103.72 $\pm$ 7.72 \\
J-24h & 3 & 269.78 $\pm$ 0.60 & 55.49 $\pm$ 0.47 & 28.15 $\pm$ 1.22 \\
\multicolumn{5}{l}{\textit{ShinkaEvolve}} \\
J1 & 3 & 279.32 $\pm$ 3.29 & 164.91 $\pm$ 4.28 & 34.73 $\pm$ 2.15 \\
J2 & 3 & 268.03 $\pm$ 1.69 & 101.39 $\pm$ 2.01 & 31.61 $\pm$ 1.95 \\
J3 & 3 & 262.38 $\pm$ 0.28 & 129.68 $\pm$ 0.47 & 33.69 $\pm$ 2.26 \\
H1 & 3 & 317.43 $\pm$ 1.55 & 62.13 $\pm$ 1.07 & 31.11 $\pm$ 6.26 \\
H2 & 3 & 329.71 $\pm$ 0.81 & 177.35 $\pm$ 1.72 & 34.52 $\pm$ 8.69 \\
J-high & 3 & 823.04 $\pm$ 8.52 & 2520.18 $\pm$ 22.89 & 115.79 $\pm$ 12.78 \\
H-high & 3 & 658.03 $\pm$ 14.62 & 1362.21 $\pm$ 53.32 & 59.34 $\pm$ 8.86 \\
J-24h & 3 & 272.42 $\pm$ 0.29 & 56.79 $\pm$ 0.20 & 26.68 $\pm$ 2.78 \\
\bottomrule
\end{tabular}

\par\smallskip\begin{minipage}{\linewidth}\scriptsize Mean $\pm$ sample SD across different fixed programs. Each program contributes one evaluation to each scenario. EvoSignal (H1, TT--QL) programs were searched on H1; the other groups were searched on J1.\end{minipage}
\end{table}

\clearpage
\section{Feature Extraction and Priority Rules of the Selected Programs}
\label{sec:selected-programs}
This appendix describes the two fixed J1 programs used in the main comparison: default EvoSignal and EvoSignal (TT--QL). The equations summarize the computations used for phase selection under the evaluated control protocol, including feature extraction, priority adjustments, and service-history updates. They assume finite, nonnegative vehicle observations and an admissible current phase. The overall phase-selection interface is defined in Section~\ref{sec:3m}. Coefficients below are constants in the selected control programs; they are distinct from the search-objective weights.

\subsection{Shared Notation and Feature Aggregation}
We follow Section~\ref{sec:3m} and omit intersection and time subscripts for a single decision. Phase $k\in P$ serves movements $B_k$, and $K=|P|=4$ in the experiments. Let $e_m$ be the observed efficient-pressure feature, $Q_m$ the waiting-vehicle count, $V_m$ the approaching-vehicle feature, and $n_m,n_m^d$ the upstream and downstream vehicle counts. Both programs use
\begin{equation}
E_k=\sum_{m\in B_k}e_m,\qquad
Q_k=\sum_{m\in B_k}Q_m,\qquad
V_k=\sum_{m\in B_k}\max(V_m,0).
\label{eq:selected-features}
\end{equation}
Thus, $E_k$ aggregates the supplied efficient-pressure feature, rather than assuming it equals a newly defined pressure formula. Let $k_0$ be the current phase after converting its observation to the program's zero-based index, and $h$ the observed \texttt{time\_this\_phase}. The simulation updates $h$ in seconds. Under the evaluation protocol, the first decision observes $h=0$. Subsequent decisions occur every 30~s: a phase change includes 5~s of yellow followed by 25~s of green, so $h\geq25$ at subsequent decision points. Retaining a phase increases its duration by another 30~s. The priority equations below describe regular decision points after initialization; initialization is stated separately. In contrast, the service-age variables below count non-transition calls to the controller; their thresholds are not durations in seconds. Define $[v]_+=\max(v,0)$ and $[v]_0^1=\min(\max(v,0),1)$.

\paragraph{Core demand terms}
For each phase, both programs combine the features defined above into a core demand term. Superscripts D and T denote default EvoSignal and EvoSignal (TT--QL), respectively:
\begin{equation}
\begin{aligned}
b_k^{\mathrm{D}}&=(1-d_k)E_k+0.8Q_k+0.4V_k,\\
b_k^{\mathrm{T}}&=E_k+(0.4+0.3\rho)Q_k+0.5V_k.
\end{aligned}
\label{eq:selected-core}
\end{equation}
The default program uses $d_k=\min(\max(D_k/15,0),1)$, where $D_k=\sum_{m\in B_k}n_m^d$ is the downstream vehicle count for the phase. The TT--QL program uses $\rho=\min(\sum_{k\in P}Q_k/(5|P|),1)$ to adjust its waiting-vehicle weight with intersection congestion. These are intermediate demand terms, rather than complete phase priorities. The following subsections specify the additional adjustments and local scoring rules.

\subsection{Default EvoSignal}
\paragraph{M1: occupancy, downstream load, and service history}
In addition to Eq.~\eqref{eq:selected-features}, M1 computes
\begin{equation}
\begin{aligned}
D_k&=\sum_{m\in B_k}n_m^d,&
O_k&=\max_{m\in B_k}(n_m/20),\\
N&=\sum_{m=1}^{12}n_m,&
C&=N/(12\times20).
\end{aligned}
\label{eq:default-features}
\end{equation}
The value 20 is a normalization constant in the program, not an independently estimated physical lane capacity. Let $u$ be the controller's counter of non-transition decision calls. With $\ell_k$ initialized to zero, M1 forms $A_k=\max(u-\ell_k,0)$. After choosing action $a$, the controller sets $\ell_a=u$. Hence $A_k$ measures calls since phase $k$ was last requested, not the accumulated waiting time of its vehicles.

\paragraph{M2: demand scoring and service-age compensation}
Using $b_k^{\mathrm{D}}$ and $d_k$ from Eq.~\eqref{eq:selected-core}, the priority before current-phase adjustments is
\begin{equation}
\widetilde q_k=
\left[b_k^{\mathrm{D}}(1+0.5[O_k]_0^1)(1+0.3C)
+0.8[A_k-4]_+(1+0.2Q_k)\right](1+0.02N).
\label{eq:default-score}
\end{equation}
The downstream factor attenuates the pressure term, upstream occupancy amplifies the demand term, and service-age compensation increases with waiting demand. At regular decision points, the local priorities are
\begin{equation}
q_k=\begin{cases}
0.4\widetilde q_k,&k=k_0,\\
0.95\widetilde q_k,&k\ne k_0.
\end{cases}
\label{eq:default-current-phase}
\end{equation}
These factors reduce the preference for retaining the current phase relative to competing phases. At initialization, the program instead sets $q_{k_0}=10^9$ and $q_k=\widetilde q_k$ for $k\ne k_0$. Signal timing follows the evaluation protocol in Section~\ref{sec:setup}.

\paragraph{M3: network-based priority adjustment}
Let $P_E$ and $P_N$ denote the queue-pressure features summed over the east--west and north--south movement groups across available shared intersection records, and let $M=\max(P_E,P_N)$. These queue-pressure summaries are distinct from the efficient-pressure feature $E_k$. Let $J_i$ count available adjacency entries whose total vehicle count exceeds the current intersection's $N$ by more than five; it is zero when adjacency or the current intersection's shared record is unavailable. The adjustment is
\begin{equation}
r_k=-0.05J_i+
\begin{cases}
0.3[P_E-P_N]_+/M,&M>1,\ k=0,\\
0.3[P_N-P_E]_+/M,&M>1,\ k=1,\\
0,&\text{otherwise}.
\end{cases}
\label{eq:default-network-adjustment}
\end{equation}
Phases 0 and 1 serve east--west and north--south through traffic, respectively. The directional term favors the through phase associated with greater network pressure, while the neighbor-load term applies the same offset to all phase scores. The code updates $g\leftarrow\max(0.1,g)$ and combines the adjustment with local priorities through Eq.~\eqref{eq:action}.

\subsection{EvoSignal (TT--QL)}
\paragraph{M1: downstream occupancy and service history}
Alongside Eq.~\eqref{eq:selected-features}, the features used by the final rule are
\begin{equation}
O_k^d=\max_{m\in B_k}(n_m^d/20),\qquad
Q_{\mathrm{tot}}=\sum_{k\in P}Q_k,\qquad
\rho=\min\left(\frac{Q_{\mathrm{tot}}}{5K},1\right).
\label{eq:ttql-features}
\end{equation}
Here $Q_{\mathrm{tot}}$ is the sum of phase waiting features used by this controller. At call $u$, M1 first sets $\ell_{k_0}=u$ for the phase currently observed active, and then computes $A_k=\max(u-\ell_k,0)$, with initial $\ell_k=0$. Unlike the default program, it updates this history from the observed current phase before scoring, rather than from the requested action after scoring. In particular, $A_{k_0}=0$.

\paragraph{M2: bounded service compensation and downstream attenuation}
Using $b_k^{\mathrm{T}}$ from Eq.~\eqref{eq:selected-core}, define the service-age bonus and the downstream factor as
\begin{equation}
\begin{aligned}
B_k^{\mathrm{age}}&=\min\left(8,\ 0.3[A_k-10]_+
+\frac{0.2[\rho-0.5]_+[A_k-10]_+}{\max(A_k,1)}\right),\\
F_k&=\begin{cases}
1,&O_k^d\leq0.8,\\
\max\left(1-2(O_k^d-0.8),0.1\right),&O_k^d>0.8.
\end{cases}
\end{aligned}
\label{eq:ttql-adjustments}
\end{equation}
The intermediate priority is
\begin{equation}
\widetilde q_k=\max\left((b_k^{\mathrm{T}}+B_k^{\mathrm{age}})F_k,0\right).
\label{eq:ttql-score}
\end{equation}
In contrast to the default program's pressure-only attenuation, $F_k$ scales the combined score. The service-age contribution is bounded by eight before this attenuation. These rules express a preference against serving movements with heavily occupied downstream links while retaining a bounded compensation for phases not recently active.

For $k\ne k_0$, let $q_k=\widetilde q_k$. Set $V_{\max}=\max_k V_k$ when any $V_k>0$, and $V_{\max}=1$ otherwise. The continuation term and current-phase priority are
\begin{equation}
\begin{aligned}
L&=0.5+0.3V_{k_0}\mathbf{1}(V_{k_0}\geq0.5V_{\max}),\\
q_{k_0}&=\begin{cases}
0.6(\widetilde q_{k_0}+L),&V_{k_0}<2,\\
\widetilde q_{k_0}+L,&\text{otherwise}.
\end{cases}
\end{aligned}
\label{eq:ttql-current-phase}
\end{equation}
The continuation bonus uses approaching demand relative to other phases. At regular decision points, the 0.6 factor reduces the current-phase priority when fewer than two approaching vehicles are observed. At initialization, the program instead sets $q_{k_0}=C_{\mathrm{clip}}$, the configured clipping bound (default $10^6$), while competing phases retain $q_k=\widetilde q_k$.

\paragraph{Phase selection}
Both programs combine scores and apply the clipping and nonfinite-value handling in Eq.~\eqref{eq:action}, then select an argmax as in Eq.~\eqref{eq:phase-selection}. The implementation resolves a tie by the first phase index. During a yellow-transition observation, the controller returns its stored action without recomputing these rules. Appendix~\ref{sec:editable-program} shows the corresponding executable interfaces, and Appendix~\ref{sec:case-output} illustrates how a recorded edit replaces part of such a program.

\subsection{Revision History of the Default Program}
\label{sec:default-revisions}
The default program develops the fairness-aware seed through substantive revisions at iterations 35, 130, and 144. The first adds efficient pressure, downstream-load attenuation, and demand terms; later edits adjust waiting and approaching weights and strengthen service-age compensation with waiting demand. Relative to the seed, the final program reduces TT, QL, and WT by 2.24\%, 5.52\%, and 7.91\%. Improvement is not monotonic: iteration 130 improves TT and QL but worsens WT and the combined score. The highest-scoring evaluation, at iteration 156, uses code identical to iteration 144.

A separate SPA event in the same search illustrates revision guided by traffic diagnostics. Its parent report identifies peak congestion in the third quarter and a 95.5\% switching rate. The revision adds efficient pressure and downstream information: TT falls from 286.51 to 280.68~s, QL from 175.17 to 167.19 vehicles, and WT from 37.20 to 34.18~s.

\clearpage
\section{An Example of Feedback-Guided Program Revision}
\label{sec:revision-example}
This appendix follows a recorded EvoSignal revision at iteration 92 on J1. The example links the actual generation input, the recorded LLM response, and the resulting evaluation. Excerpts retain the recorded wording; \texttt{[...]} marks omitted material. Line wrapping, indentation, and typographic punctuation are normalized for readability. Explanatory text outside the boxes is supplied by the authors.

\subsection{Editable Program Structure}
\label{sec:editable-program}
The following structural excerpt shows how the controller invokes its three editable modules. Dots omit module bodies, transition handling, shared-feature publication, diagnostics, and program-specific history updates. The names match the implementation. Evolution revises the computations within M1--M3 while retaining their interfaces and the prescribed phase-selection operation. The functions may update controller state, including the M3 gate. Detailed feature extraction and priority rules for the two selected programs are given in Appendix~\ref{sec:selected-programs}.
\begin{lstlisting}[style=caseexcerpt,language=Python]
def choose_action(self, count, state):
    # Transition handling and counter update omitted.
    local_feats, global_feats = self.feature_extractor(state)
    # Publish local features to the shared store.
    priority_local = self.local_priority(local_feats)
    priority_delta = self.global_modifier(global_feats, local_feats)
    priority_final = priority_local + self.global_gate * priority_delta
    priority_final = np.clip(priority_final,
                             -self.priority_clip, self.priority_clip)
    priority_final = np.nan_to_num(priority_final, nan=0.0,
                                   posinf=0.0, neginf=0.0)
    self.action = int(np.argmax(priority_final))
    # History updates omitted.
    return self.action

# EVOLVE-BLOCK-START
def feature_extractor(self, state):                 # M1
    ...  # Extract traffic features and context.
    return local_feats, global_feats

def local_priority(self, local_feats):              # M2
    ...  # Compute one base priority per phase.
    return priority

def global_modifier(self, global_feats, local_feats):  # M3
    ...  # Compute one adjustment per phase.
    return delta
# EVOLVE-BLOCK-END
\end{lstlisting}
Feature aggregation and history construction can change in M1; coefficients, nonlinear expressions, and conditional phase-scoring rules can change in M2; network-context corrections can change in M3. These are program-design variables, rather than fixed parameters tuned within a single scoring formula.

\subsection{Input: Task and Traffic Diagnostics}
\label{sec:case-spa}
Table~\ref{tab:prompt-map} maps the prompt components to the method notation. The full input also includes parent code and reference programs; the following excerpts focus on task guidance, SPA, and PSA.
\begin{table}[H]
\centering\small
\caption{Components of the recorded revision prompt.}
\label{tab:prompt-map}
\begin{tabular}{@{}p{0.26\linewidth}p{0.69\linewidth}@{}}
\toprule
Input component & Role in program revision \\
\midrule
Task and interfaces ($G_n$) & Define the control task, editable modules, and current M2 focus. \\
Parent and references ($p_a,R_n$) & Supply executable code and alternative control examples. \\
Performance and SPA ($m_a$) & Report parent metrics, congestion patterns, and suggested changes. \\
PSA summary ($b_a$ in $m_a$) & Describe retained strategies and their measured trade-offs. \\
Output instructions & Request explanations followed by exact SEARCH/REPLACE edits. \\
\bottomrule
\end{tabular}
\end{table}
\par\medskip\noindent\textbf{System prompt: control task and revision focus}\par
\begin{lstlisting}[style=caseexcerpt]
You are an expert traffic signal control engineer and algorithm designer.
Your task is to improve a traffic signal controller called
PriorityAgent that runs inside the CityFlow traffic simulator.
The algorithm must be **general-purpose** -- it should work across different
road networks and traffic demand patterns, not overfit to one scenario.
[...]
## Current Focus: PRIORITY LOGIC REFINEMENT (M2 focus)

M1 should already provide useful features.  **Focus primarily on M2
(local_priority)** -- refine how features are combined into priority.
[...]
- Keep M3 conservative unless you are confident coordination will help.
\end{lstlisting}
\par\medskip\noindent\textbf{SPA: parent performance and congestion feedback}\par
\begin{lstlisting}[style=caseexcerpt]
[Overall] avg_tt=277.7s | avg_queue=163.9 | avg_wait=36.5s | reward=0.0 | steps=120 | global_gate=0.0
[...]
[Temporal Trend]
  Q1(0-900s)=129.4 -> Q2(900-1800s)=145.7 -> Q3(1800-2700s)=199.0 -> Q4(2700-3600s)=175.3
  Peak congestion at Q3 (1800-2700s)
[...]
## Mutation Suggestions (from diagnostic analysis)

1. Congestion peaks in Q3 (1800-2700s), suggesting the algorithm struggles under sustained load. Consider adaptive mechanisms: increase priority sensitivity when total queue grows, or add a time-varying urgency multiplier.
\end{lstlisting}

\clearpage
\subsection{Input: Strategy Memory; Output: Revision Rationale}
\label{sec:case-psa}
The PSA excerpt below is from the summary actually supplied to this generation. It describes an earlier strategy whose score exceeded the parent's current score. The displayed \texttt{iter -1} is a missing iteration identifier in the saved summary, not the iteration of the current revision. The archive heading and diagnostic suggestions are reproduced as recorded.
\par\medskip\noindent\textbf{PSA: one retained strategy}\par
\begin{lstlisting}[style=caseexcerpt]
## Strategy Archive (71 evaluated, 8 on Pareto front)

### Pareto Front (non-dominated solutions)
Study their module designs to understand trade-offs.

1. [iter -1] TT=276.4s  QL=161.7  WT=35.6s  (combined=0.3961)
   Strengths: none
   M1: Computes per-phase {phase_pressure_queue, phase_pressure_queue_efficient, phase_vehicle_count, phase_waiting, phase_running, phase_vc_downstream, phase_time_since_green} plus per-lane {lane_occupancy} and temporal {time_this_phase}.
   M2: Computes local priority per phase as base = ph_pqe + 0.4*ph_wt + 0.1*ph_run - 0.15*ph_vc_down + 0.8*max_lane_occ; adds starvation bonus if ph_tsg > STARVATION_THRESHOLD (4): 0.5*(ph_tsg-4)*(1+0.1*ph_wt); reduces current phase priority by factor 0.3 if ttp >= MAX_GREEN_STEPS (6).
   [...]
\end{lstlisting}

The response explicitly refers to the archive and proposes simplifying M1 and restoring a local-priority rule. These are the model's recorded explanations; the code and measured outcomes are examined separately on the next page. Duration labels in the response retain the generated wording; the controller receives elapsed phase time in seconds and tests it only at scheduled decision points.
\par\medskip\noindent\textbf{LLM response: stated revision and planned edits}\par
\begin{lstlisting}[style=caseexcerpt]
We need to revert local_priority to the simpler, proven formula from the Pareto front -- efficient pressure + fixed 0.4 * waiting + 0.1 * running - 0.15 * downstream_count + 0.8 * maximum lane occupancy per phase, plus a linear anti-starvation bonus for phases starved > 4 steps, and only a max-green safeguard (reduce priority by 0.3 after 6 steps). This matches the best-performing agents and removes all the unnecessary complexity (adaptive weight, spillback penalty, delta-waiting reward, urgency multiplier, min-green override) that degraded fitness from 0.3961 to 0.3930.

We also simplify M1 by eliminating the temporal derivative (ph_wt_delta, prev_ph_wt), total_waiting, overall_occupancy, step_urgency, and downstream occupancy features, keeping only the essential phase-level aggregates that M2 actually uses and that are required by the network-summary diagnostic.

Three targeted replacements:

1. **M1 aggregation block** - drop all extra computations.
2. **M1 local_feats dict** - keep only the keys needed by M2 and shared diagnostics.
3. **Entire M2 local_priority method** - replace with the simple, robust formula.
\end{lstlisting}

SPA suggests congestion-sensitive adaptation, while PSA supplies a simpler alternative with a stronger recorded score. In this event, the response chooses the latter direction. This illustrates how diagnostic feedback and retained strategies provide context for revision; the LLM is not required to implement every suggestion.

\clearpage
\subsection{Output: Code Revision and Measured Outcome}
\label{sec:case-output}
The response contains three SEARCH/REPLACE edits: two simplify M1 feature construction and one replaces M2. Below are excerpts from the original M2 and the replacement. The replacement is present in both the response and the saved child program. For compactness, the surrounding method definitions, feature retrieval, and other unchanged code are omitted.
\par\medskip\noindent\textbf{Recorded SEARCH/REPLACE format (M2 excerpt)}\par
\begin{lstlisting}[style=caseexcerpt]
    <<<<<<< SEARCH
    def local_priority(self, local_feats):
        [...]
            base = (ph_pqe[p] + w_wait * ph_wt[p] + 0.1 * ph_run[p]
                    + 0.8 * ph_max_occ[p])
        [...]
    =======
    def local_priority(self, local_feats):
        [...]
            base = ph_pqe[p] + 0.4 * ph_wt[p] + 0.1 * ph_run[p] - 0.15 * ph_vc_down[p]
        [...]
    >>>>>>> REPLACE
\end{lstlisting}
The SEARCH block identifies the original method text, and the REPLACE block supplies its replacement. The recorded edit contains the complete method on each side; the displayed \texttt{[...]} omissions are for presentation and are not part of the executable edit. The excerpts below explain the changed computation. This intermediate revision illustrates the editing mechanism; it is distinct from the two final programs analyzed in Appendix~\ref{sec:selected-programs}.

\par\medskip\noindent\textbf{Before: congestion-dependent weighting}\par
\begin{lstlisting}[style=caseexcerpt,language=Python]
congestion_factor = min(1.0, total_waiting / 50.0)   # 0-1
w_wait = 0.4 + 0.4 * congestion_factor               # 0.4 -> 0.8
[...]
base = (ph_pqe[p] + w_wait * ph_wt[p] + 0.1 * ph_run[p]
        + 0.8 * ph_max_occ[p])
\end{lstlisting}
\par\medskip\noindent\textbf{After: the replacement priority computation}\par
\begin{lstlisting}[style=caseexcerpt,language=Python]
for p in range(n):
    base = ph_pqe[p] + 0.4 * ph_wt[p] + 0.1 * ph_run[p] - 0.15 * ph_vc_down[p]

    max_occ = 0.0
    for m in self.phase_movements.get(p, []):
        if m < len(lane_occ):
            max_occ = max(max_occ, lane_occ[m])
    base += 0.8 * max_occ

    if ph_tsg[p] > STARVATION_THRESHOLD:
        starvation_bonus = 0.5 * (ph_tsg[p] - STARVATION_THRESHOLD) * (1.0 + ph_wt[p] * 0.1)
        base += starvation_bonus

    priority[p] = base

cur_0idx = cur - 1 if cur > 0 else 0
if 0 <= cur_0idx < n and ttp >= MAX_GREEN_STEPS:
    priority[cur_0idx] *= 0.3
\end{lstlisting}

Here \texttt{ph\_pqe}, \texttt{ph\_wt}, \texttt{ph\_run}, and \texttt{ph\_vc\_down} represent efficient pressure, waiting vehicles, approaching vehicles, and downstream vehicle counts. The code adds occupancy and service-age terms, then adjusts the current phase's priority. The recorded constants are \texttt{STARVATION\_THRESHOLD = 4} and \texttt{MAX\_GREEN\_STEPS = 6}.

\begin{table}[H]
\centering\small
\caption{Recorded parent and child performance for the illustrated revision.}
\label{tab:revision-outcome}
\begin{tabular}{lrr}
\toprule
Metric & Parent & Revised program \\
\midrule
TT (s) & 277.70 & 276.37 \\
QL (veh) & 163.89 & 161.73 \\
WT (s) & 36.52 & 35.64 \\
Search score & 0.392953 & 0.396055 \\
\bottomrule
\end{tabular}
\end{table}
All three traffic metrics improve relative to the parent. The resulting score recovers the level of an earlier archived strategy. This is a documented example of strategy reuse and program revision, rather than an isolated estimate of SPA or PSA's causal contribution.

\clearpage
\section{Objective-Weight Analysis and Additional Views}
\label{sec:pareto-full}
\paragraph{Score contributions and candidate distributions}
For the two selected J1 programs, scoring under the default weights gives a combined TT and QL gain of about 0.00514 but a WT cost of 0.01483 when moving from default EvoSignal to TT--QL. Under the TT--QL weights, the gain becomes 0.00755 and the cost falls to 0.00198, reversing their ranking. Each comparison applies one common objective to both programs. Among evolved candidates, median TT decreases from 279.22 to 274.22~s and median QL from 165.06 to 159.77 vehicles under the TT--QL objective, while median WT increases from 35.34 to 43.86~s. These statistics describe the two recorded searches.

\begin{figure}[!htbp]
\centering
\includegraphics[width=\linewidth]{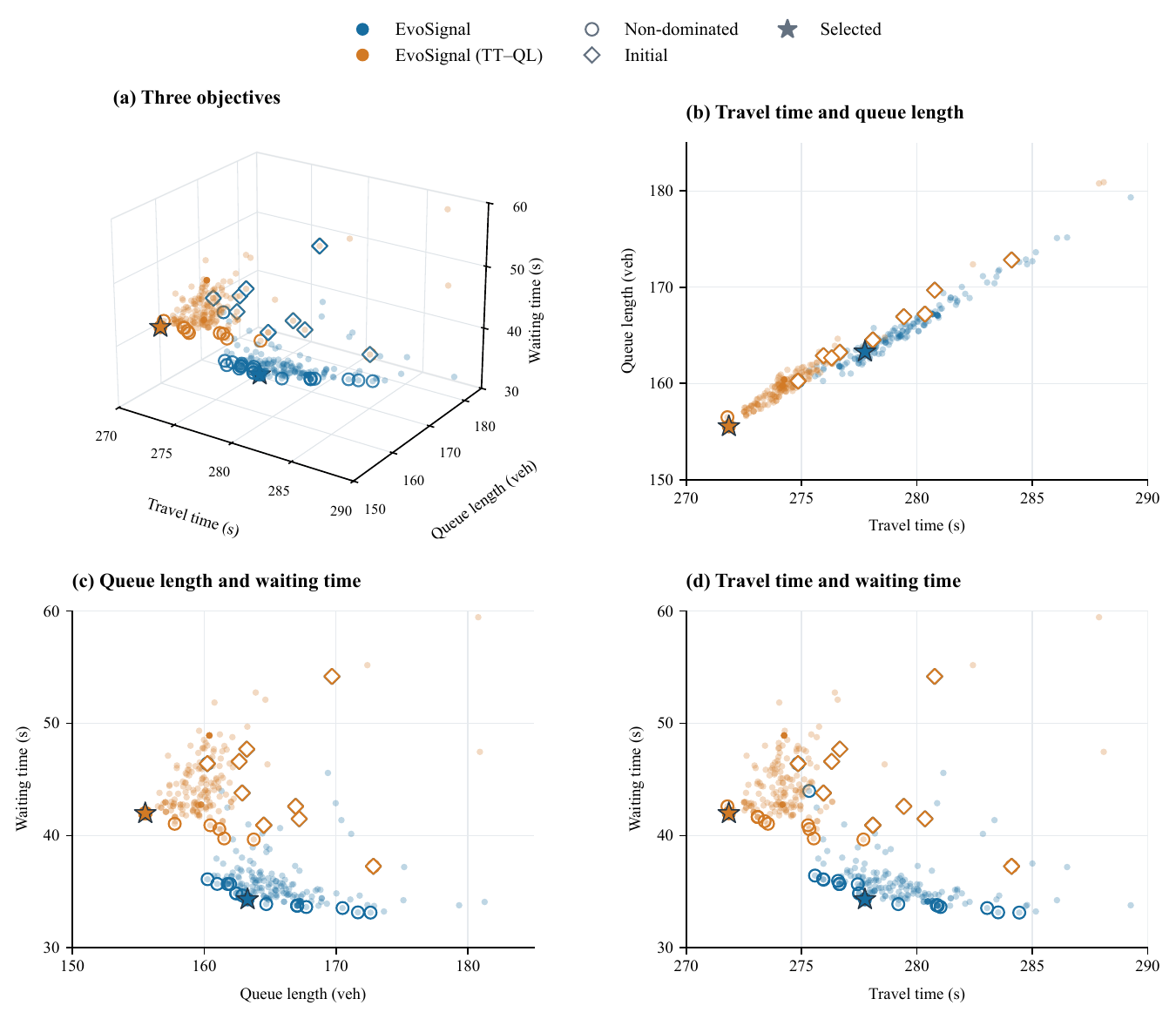}
\caption{Additional objective-space views of the default and TT--QL searches on J1: (a) three objectives; (b--d) pairwise projections over wider display ranges.}
\label{fig:objective-pareto-full}
\end{figure}

\paragraph{Records and display ranges}
The EvoSignal and TT--QL searches provide 197 and 200 valid records, respectively. Each search includes 11 initial programs, which enter the plots and non-dominated sets; the reported medians exclude initialization. A program is non-dominated if no other evaluated program in the same search is no worse on every displayed metric and strictly better on at least one. Fronts are computed jointly for the three metrics in the three-dimensional view and independently for each metric pair in the two-dimensional views. This does not establish global Pareto optimality.

Figure~\ref{fig:objective-pareto} uses TT of 270.5--286~s, QL of 153--176 vehicles, and WT of 32--51~s; its two panels show 187/188 EvoSignal records and 195/190 TT--QL records. Figure~\ref{fig:objective-pareto-full} expands these limits to 270--290~s, 150--185 vehicles, and 30--60~s, respectively. Panels (a)--(d) show 190/190/191/190 EvoSignal records and 197 TT--QL records each. All empirical non-dominated points and both selected programs remain visible; out-of-range records remain in the front calculations.

\FloatBarrier
\section{Implementation Details and Hyperparameters}
\label{sec:implementation-settings}
Table~\ref{tab:evolution-settings} summarizes the core settings from the archived main EvoSignal runs. The objective weights for the three evaluated configurations are defined in Section~\ref{sec:evosignal-variants}. The full framework enables 3M, Multi-init, SPA, and PSA; each ablation removes the corresponding component.

\begin{table}[!htbp]
\centering\small
\caption{Core evolution and traffic-control settings.}
\label{tab:evolution-settings}
\renewcommand{\arraystretch}{1.13}
\begin{tabular}{@{}p{0.52\linewidth}p{0.43\linewidth}@{}}
\toprule
Setting & Value \\
\midrule
\multicolumn{2}{@{}l}{\textit{Program evolution and LLM generation}} \\
Evolution attempts after initialization, $N$ & 200 \\
Initial programs (full / without Multi-init) & 11 / 1 \\
LLM sampling probabilities (Flash / Pro) & 0.7 / 0.3 \\
Generation temperature / maximum output tokens & 0.7 / 16,000 \\
Population size / database archive size / islands & 40 / 15 / 2 \\
Elite-selection / exploitation ratios & 0.2 / 0.6 \\
Top / diverse reference programs in the prompt & 2 / 1 \\
Concurrent evaluations / evaluation timeout & 2 / 600~s \\
Code revision format & SEARCH/REPLACE within editable blocks \\
\midrule
\multicolumn{2}{@{}l}{\textit{Objective scaling and strategy memory}} \\
Score scales $(C_T,C_Q,C_W)$ & (100~s, 100 vehicles, 50~s) \\
PSA memory capacity, $L$ & 50 records \\
Maximum front / recent dominated entries shown & 8 / 5 \\
\midrule
\multicolumn{2}{@{}l}{\textit{Traffic simulation and signal control}} \\
Simulation step / phase-decision interval & 1~s / 30~s \\
Yellow transition when changing phase & 5~s within the decision interval \\
Selectable phases per intersection & 4 \\
Evaluation horizon & 3,600~s; 86,400~s for J-24h \\
\bottomrule
\end{tabular}
\end{table}

Flash and Pro denote DeepSeek-V4-Flash and DeepSeek-V4-Pro. The database archive stores programs; the separate PSA memory retains strategy records, with limits on the front and recent entries supplied to the LLM.

The pressure--demand program is both a member of the full initial set and the sole seed for the without-Multi-init ablation.

\subsection{Initial Programs and Feature Examples}
\label{sec:initial-program-details}
Table~\ref{tab:initial-strategies} lists the 11 experimental seeds. They were generated with LLM assistance from complementary control ideas, implemented through the same 3M interfaces, and tested for successful execution. Their names indicate design inspiration rather than complete reproductions of classical controllers.
\begin{table}[!htbp]
\centering\small
\caption{Initial control programs and their complementary design principles.}
\label{tab:initial-strategies}
\setlength{\tabcolsep}{4pt}
\renewcommand{\arraystretch}{1.12}
\begin{tabular}{@{}>{\raggedright\arraybackslash}p{0.25\linewidth}>{\raggedright\arraybackslash}p{0.22\linewidth}>{\raggedright\arraybackslash}p{0.47\linewidth}@{}}
\toprule
Initial program & Design emphasis & Main control idea \\
\midrule
Pressure--demand & Pressure and demand & Combine queue pressure, waiting vehicles, and approaching vehicles to rank phases. \\
\addlinespace[3pt]
Efficient pressure & Pressure-based control & Use efficient queue pressure directly as a simple initial priority rule. \\
\addlinespace[3pt]
Continuation-aware pressure & Phase continuation & Favor the current phase when approaching demand justifies continuation; otherwise follow pressure priorities. \\
\addlinespace[3pt]
Demand--supply balance & Demand and capacity & Combine estimated demand--supply ratios, saturation, and waiting vehicles to assess service urgency. \\
\addlinespace[3pt]
PID-inspired & Feedback control & Combine pressure, an accumulated-waiting proxy, and queue growth to respond to current and past congestion. \\
\addlinespace[3pt]
Fairness-aware & Service fairness & Increase priorities for phases not recently selected and reduce preference for a phase retained for a long period. \\
\addlinespace[3pt]
Corridor coordination & Neighbor coordination & Adjust local priorities using neighboring traffic information to encourage coordinated service. \\
\addlinespace[3pt]
Webster-inspired & Capacity-based allocation & Use estimated volume-to-capacity ratios and waiting demand to determine phase priorities. \\
\addlinespace[3pt]
Lane balancing & Lane congestion & Prioritize congested lanes while penalizing movements with high downstream occupancy. \\
\addlinespace[3pt]
Intersection adaptation & Congestion adaptation & Adjust the relative contributions of pressure and waiting vehicles according to intersection congestion. \\
\addlinespace[3pt]
Network trends & Network coordination & Modify local priorities according to dominant directional demand across the network. \\
\bottomrule
\end{tabular}
\par\smallskip\begin{minipage}{\linewidth}\scriptsize Names indicate design inspiration; the programs implement these ideas within the 3M phase-priority structure.\end{minipage}
\end{table}

\paragraph{M1: Traffic feature extraction}
M1 transforms movement-level observations into information useful for phase selection. For example, in the base initial program, let $B_k$ denote the movements served by phase $k$. For movement $m$, let $\pi_m$ denote the observed queue-pressure feature, $Q_m$ the waiting-vehicle count, and $V_m$ the moving-vehicle demand feature. M1 aggregates them as
\begin{equation}
\pi_k=\sum_{m\in B_k}\pi_m,\qquad
Q_k=\sum_{m\in B_k}Q_m,\qquad
V_k=\sum_{m\in B_k}\max(V_m,0).
\label{eq:phase-features}
\end{equation}
These phase features are included in $z$, alongside movement-level information and the current signal state. M1 can be revised to introduce downstream congestion, demand changes, service history, or different feature combinations. The program also publishes features to a shared intersection store. Since intersections update sequentially, shared entries may come from different decision steps. Context $c$ can include adjacency information, shared features, and network summaries.

\paragraph{M2: Local phase prioritization}
M2 assigns each phase a base score $q_k$. In the same initial program, this is a weighted combination of the features in Eq.~\eqref{eq:phase-features}:
\begin{equation}
q_k=\alpha\pi_k+\beta Q_k+\gamma V_k,
\label{eq:seed-priority}
\end{equation}
where $\alpha$, $\beta$, and $\gamma$ are control-rule coefficients. They are distinct from the search-objective weights $w_j$ in Eq.~\eqref{eq:fitness}. This example shows how pressure, waiting traffic, and approaching demand can influence phase priority. Evolution can revise the coefficients and replace the linear expression with nonlinear, conditional, or history-dependent rules. Thus, Eq.~\eqref{eq:seed-priority} illustrates one starting rule rather than restricting all candidates to a fixed functional form.
The pressure--demand seed uses mean phase-pressure features from available neighbors for its network-adjustment rule. Setting its initial gate to zero removes this additive effect; the scaffold still computes $r$.

\subsection{Database Records and Evolution Updates}
\label{sec:evolution-details}
For each successfully evaluated program, the database stores
\begin{equation}
m_j=\bigl(\mathbf{f}(p_j),S(p_j),d_j,b_j\bigr),
\label{eq:revision-record}
\end{equation}
Here $d_j$ and $b_j$ are SPA feedback and the saved PSA summary; unavailable feedback is omitted. Code, execution status, and parent identity are also retained. Selection uses the inherited island-based database, combining elite retention with exploratory and random sampling:
\begin{equation}
(p_a,R_n)=\operatorname{Select}(D_{n-1}),
\label{eq:parent-selection}
\end{equation}
References include high-scoring and alternative programs. A successful evaluation returns
\begin{equation}
\bigl(\mathbf{f}(p_n),\tau_n\bigr)=E(p_n),
\label{eq:traffic-evaluation}
\end{equation}
where $E$ uses the specified network, demand, and horizon. After forming $m_n$, the database update is
\begin{equation}
D_n=U\bigl(D_{n-1},(p_n,m_n),a\bigr),
\label{eq:revision-update}
\end{equation}
The operator $U$ stores the candidate and parent identity $a$ under the database retention rules. Construction and evaluation failures retain separate status records. Workers can evaluate concurrently, so updates follow completion order.

Successful seed evaluations initialize the database through
\begin{equation}
m^{(k)}=\bigl(\mathbf{f}(p^{(k)}),S(p^{(k)}),\varnothing,\varnothing\bigr),
\qquad
D_0=\operatorname{Init}\bigl(\{(p^{(k)},m^{(k)})\}_{k=1}^{K}\bigr).
\label{eq:initial-database}
\end{equation}
Here $\operatorname{Init}$ inserts evaluated seeds under the same retention rules. Seeds can update PSA, but the initialization path does not save SPA reports or PSA snapshots in their database records; the empty fields reflect that implementation. Failed seed evaluations retain their execution status.

\subsection{Additional SPA Statistics and Diagnostic Rules}
\label{sec:diagnostic-details}
Using the notation in Section~\ref{sec:spa}, SPA computes the switching rate for $T>1$ as
\begin{equation}
v_{\mathrm{sw}}=\frac{1}{I(T-1)}\sum_{i=1}^{I}\sum_{t=2}^{T}
\mathbf{1}(a_{i,t}\ne a_{i,t-1}).
\label{eq:spa-switching}
\end{equation}
The mean queue at intersection $i$ and the mean network queue in temporal window $W$ are
\begin{equation}
\overline{Q}_i=\frac{1}{T}\sum_{t=1}^{T}\sum_m Q_{i,t,m},
\qquad
\overline{Q}_{W}=\frac{1}{|W|}\sum_{t\in W}\sum_{i=1}^{I}\sum_m Q_{i,t,m}.
\label{eq:spa-queues}
\end{equation}
The implementation compares four consecutive temporal windows and groups intersections by location. It also examines phase-score variation, demand--priority agreement, and observations following module changes. These statistics guide inspection rather than estimate causal effects. The report contains up to four revision suggestions. For example, $f_W(p_n)>1.5C_W$ prompts examination of green-hold or switching-margin rules; other rules address uneven service, demand alignment, and late congestion.

\subsection{PSA Storage and Retrieval}
\label{sec:archive-details}
With module summaries $s_n$ defined in Section~\ref{sec:psa}, the archive record and memory update are
\begin{equation}
\begin{aligned}
e_n&=\bigl(\mathbf{f}(p_n),\operatorname{round}(S(p_n),4),s_n\bigr),\\
M^+&=\operatorname{Tail}_{L}(M\mathbin{\Vert}e_n).
\end{aligned}
\label{eq:psa-update}
\end{equation}
Here $\Vert$ appends a record and $\operatorname{Tail}_L$ retains the latest $L$ records, with $L=50$. Records are linked to program identities, stored scores are rounded to four decimal places, and summaries are cached when identical code reappears. Successful seed evaluations also enter this memory. Given the nondominated set $F$, the saved text summary is
\begin{equation}
b_n=\operatorname{Render}\bigl(\operatorname{Top}_{8}(F),
\operatorname{Recent}_{5}(M^+\setminus F)\bigr).
\label{eq:psa-summary}
\end{equation}
The first operator selects up to eight nondominated records by stored scalar score in descending order. The second selects up to five recent dominated records, newest first. Rendering combines their metrics and module summaries. The summary is stored with its candidate, so a later parent revision uses that saved snapshot rather than necessarily the latest memory state.

\end{document}